\documentclass[journal]{IEEEtran}
\ifCLASSINFOpdf
\else
\fi

\usepackage{amsmath}
\usepackage{times}
\usepackage{graphicx}
\usepackage{color}
\usepackage{multirow}
\usepackage{amsmath,amssymb,nccmath,mathtools}
\usepackage{rotating}
\usepackage{bbm}
\usepackage{latexsym}
\usepackage{hyperref}

\usepackage{caption}
\begin{document}

\def \xx {\mathbf{x}}
\def \ww {\mathbf{w}}
\def \yy {\mathbf{y}}
\def \bb {\mathbf{b}}
\def \R {\mathbb{R}}

\title{Expert2Coder: Capturing Divergent Brain Regions Using Mixture of Regression Experts}

\author{Subba~Reddy~Oota,~\IEEEmembership{Member,~IEEE,}
        Naresh~Manwani,~\IEEEmembership{Member,~IEEE,}
        Raju~S.~Bapi,~\IEEEmembership{Senior~Member,~IEEE}% <-this % stops a space
\thanks{Subba Reddy Oota is with the Cognitive Science Lab (CSL) and Machine Learning Lab (MLL),
Kohli Center on Intelligent Systems (KCIS), International Institute of Information Technology (IIIT), Hyderabad,
India, 500032 e-mail: oota.subba@students.iiit.ac.in.}% <-this % stops a space
\thanks{Naresh Manwani and Raju S. Bapi are, respectively, with the Machine Learning Lab (MLL) and Cognitive Science Lab (CSL),
Kohli Center on Intelligent Systems (KCIS), International Institut  of Information Technology (IIIT), Hyderabad, India, 500032 e-mail: naresh.manwani@iiit.ac.in, raju.bapi@iiit.ac.in.}% <-this % stops a space
}

\maketitle

\begin{abstract}
fMRI semantic category understanding using linguistic encoding models attempts to learn a forward mapping that relates stimuli to the corresponding brain activation. 
State-of-the-art encoding models use a single global model (linear or non-linear) to predict brain activation (all the voxels) given the stimulus. %early detection of activated regions for unknown stimuli is challenging.
However, the critical assumption in these methods is that \emph{a priori} different brain regions respond the same way to all the stimuli, that is, there is no modularity or specialization assumed for any region. This goes against the modularity theory, supported by many cognitive neuroscience investigations suggesting that there are functionally specialized regions in the brain.
In this paper we achieve this by clustering similar regions together and for every cluster we learn a different linear regression model using a mixture of linear experts model. The key idea here is that each linear expert captures the behaviour of similar brain regions.
%In this paper, we present a mixture of experts model for predicting brain activity patterns. 
Given a new stimulus, the utility of the proposed model is twofold (i) predicts the brain activation as a weighted linear combination of the activations of multiple linear experts and (ii) to learn multiple experts corresponding to different brain regions. We argue that each expert captures activity patterns related to a particular region of interest (ROI) in the human brain. 
%Thus, the utility of the proposed model is twofold. 
%It not only accurately predicts the brain activation for a given stimulus, but it also reveals the level of activation of individual brain regions. 
%The results of our experiments highlight the importance of the proposed model for predicting brain activation.
%We also showcase that the proposed mixture of experts-based model indeed maps experts to brain regions based on the activation patterns.
This study helps in understanding the brain regions that are activated together given different kinds of stimuli. Importantly, we suggest that the mixture of regression experts (MoRE) framework successfully combines the two principles of organization of function in the brain, namely that of \emph{specialization} and \emph{integration}.
Experiments on fMRI data from paradigm 1~\cite{pereira2018toward} where participants view linguistic stimuli show that the proposed MoRE model has better prediction accuracy  compared to that of conventional models. 
Our model achieves a mean absolute error (MAE) of 3.94, with an $R^2$-score of 0.45  on this data set. This is an improvement over performance of traditional methods including,
ridge regression (5.58 MAE, 0.15 $R^2$-score), MLP (4.63 MAE, 0.35 $R^2$-score). We also elaborate on the specializations captured by various experts in our mixture model and their implications.
\end{abstract}


% \fi
\IEEEpeerreviewmaketitle

\section{Introduction}
Functional magnetic resonance imaging (fMRI) measures brain activity by identifying the changes in the blood-oxygen-level-dependent (BOLD) imaging signals in different functional areas in response to particular stimuli.
In recent years, the use of both linear and non-linear multivariate encoding or decoding approaches for analyzing fMRI brain activity has become increasingly popular \cite{mitchell2008predicting,naselaris2011encoding,pereira2018toward,hoefle2018identifying,du2018reconstructing,du2019brain}.
An encoding model that predicts brain activity in response to stimuli is essential for the neuroscience community as the model predictions are useful to investigate and test hypotheses about the transformation from stimulus to brain responses both in the healthy brain and their breakdown in clinical conditions~\cite{paninski2007statistical, kay2008identifying, dumoulin2008population, yamins2014performance, gucclu2017increasingly}.
Typically, experimental conditions involve sensory (visual or auditory), or motor stimuli, so an encoding model maps the input stimuli to their encoding representation in the respective brain region~\cite{kay2008identifying, mitchell2008predicting, dumoulin2008population, pereira2018toward, subba2019StepEncog}.

Linguistic stimuli such as words/sentences are extensively used in fMRI experiments. Understanding the association between the semantics of words/sentences and evoked brain activation may throw light on how the brain organizes and represents linguistic information in neural circuits. One of the pioneering works of Mitchell et al.~\cite{mitchell2004learning,mitchell2008predicting} proposed distributional semantic models that encode patterns found in fMRI brain activation based on hand-designed features.
Subsequently, models trained using word embedding have successfully overcome the limitations of manually-designed features to build encoding systems \cite{oota2018fmri,abnar2018experiential,pereira2018toward}. Psycholinguistic and behavioral characteristics are also useful for the encoding task \cite{chang2011quantitative,palatucci2009zero,fernandino2015predicting} and for visually-grounded representations~\cite{anderson2017visually}.
These studies establish a higher correlation between the semantic features and brain activation patterns. However, they do not have a principled way to predict regions that specialize in a particular category of stimuli. 
Instead, they predict the voxel intensity values for the whole brain or a pre-selected set of voxels.

Classical encoding models focus on univariate fMRI analysis, i.e., toward an understanding of different cognitive processes at individual brain voxels~\cite{gonsalves2010brain}. 
Researchers have also explored multi-voxel pattern analysis (MVPA)~\cite{mahmoudi2012multivoxel} to represent information across an ensemble of voxels.
The critical limitation of MVPA is that it may detect areas where brain activation differs across subjects, even if those differences are unrelated to neural coding.
Moreover, current encoding methods attempt to learn either weights in case of linear models~\cite{mitchell2008predicting,pereira2018toward} or complex representations in non-linear models~\cite{oota2018fmri}.
Recent studies show that deep learning models (e.g., convolution neural networks and LSTMs) are successful in encoding brain responses for various sensory inputs (auditory, visual, and linguistic)~\cite{wen2018transferring,han2018variational,rowtula2018deep,wen2017neural, subba2019StepEncog}. 
However, it remains unclear to what extent the deep learning models can exhibit principles of integration and differentiation that are the hallmark of how the brain responds to sensory stimulation. Moreover, all these models vary in their complexity.  In particular, interpretation of these non-linear models can be difficult due to unexpected and enigmatic representations \cite{benjamin2017modern,kording2018roles}. 

It is often incongruous to get a single global model achieving the best results on the complete problem domain~\cite{dietterich2000ensemble}. Two fundamental principles of organization of function in the brain seem to be \emph{functional differentiation} (specialization) and \emph{functional integration} (Friston, 2002). Extant linear and non-linear models tend to conform to the latter principle by modeling the integrative aspect in a single global model. However, we hypothesize that such integration is mediated in turn by regional functional specialization. Such a framework posits that information organization in the brain is achieved holistically by combining the principles of differentiation and integration. A machine learning framework that elegantly combines these principles is the \emph{mixture of experts} model~\cite{jordan1995convergence}.
% As a result, using multiple local regression models to reduce the error and also easy to interpret the overall regression system. Moreover, if one of the regressor fails still we can achieve the overall system error~\cite{kotsiantis2011combining}.

Mixture of Experts (MoE) models \cite{jordan1995convergence,yuksel2012twenty} offer an exciting choice for the problem of learning distinct models for different regions of the input space. In mixture of experts, the feature space is probabilistically divided into several partitions. Every expert specializes over a distinct partition. MoE has been used successfully to investigate the intricate patterns of brain changes associated with non-pathological and pathological processes, such as the effects of growth, aging, injury, or a  disease~\cite{kim2010bayesian,eavani2016capturing}. 
Models involving MoEs have great potential for use in medical diagnostics to diagnose a variety of clinical conditions such as depression, Alzheimers' dementia. 
Yao et al.~\cite{yao2009hierarchical} proposed a hidden conditional random field (HCRF) framework in combination with a mixture of experts model to make predictions in all ROIs that are interconnected.

In this paper, we use the MoE model assuming that each expert specializes over a particular brain region (set of voxels that are significantly activated together) based on the category of words that are represented by the model. 
Encoding models have proven successful in using pre-trained word embedding methods such as Word2Vec and GloVe to predict brain responses~\cite{mikolov2013distributed,pennington2014glove}.
Here, we use bidirectional encoder representations from transformers (BERT) embeddings~\cite{devlin2018bert} to generate a feature vector for input stimuli.
The stimuli used in task-specific fMRI datasets arise from multiple categories of data and yield activation in different brain regions. The main objective is to demonstrate the feasibility of extracting brain activity patterns related to particular regions of specialization using a mixture of regression experts-based model (we call, {\bf Expert2Coder}) while maintaining comparable accuracy of that of integrative global models.
%Finally, we modeled the fMRI signals as a linear combination of all expert predictions corresponding to the gate probabilities of each expert.
We conducted simulation experiments demonstrating that functional differentiation into divergent brain regions is indeed achieved with the mixture of regression experts model rather than using a simple linear/non-linear model alone. In summary, we make the following contributions in this paper.
\begin{enumerate}
    \item We propose a mixture of experts based model in which brain activity patterns related to each region of interest (ROI) is represented using a group of experts.
    \item In particular, we focus on categorizing different brain regions associated with different experts, given input stimuli.
    \item Showcase and highlight the importance of discrimination across different experts.
    \item Better the accuracy of the proposed model to that achieved by existing linear and non-linear models, thereby demonstrating the functional integration capability of such mixture models.
\end{enumerate}
 
The rest of this paper is organized as follows. We discuss the proposed mixture of regression experts (MoRE) approach in  Section~\ref{Approach}, and our enhancements include a detailed analysis of the dataset, insights from analysis of results and discussion in Sections~\ref{headings} and~\ref{sec:results}. Finally, we conclude with a summary in Section~\ref{sec:conclusion}.

\section{Proposed Approach: Expert2Coder}
\label{Approach}
\begin{figure*}[t]
    \centering
    \includegraphics[width=0.9\linewidth]{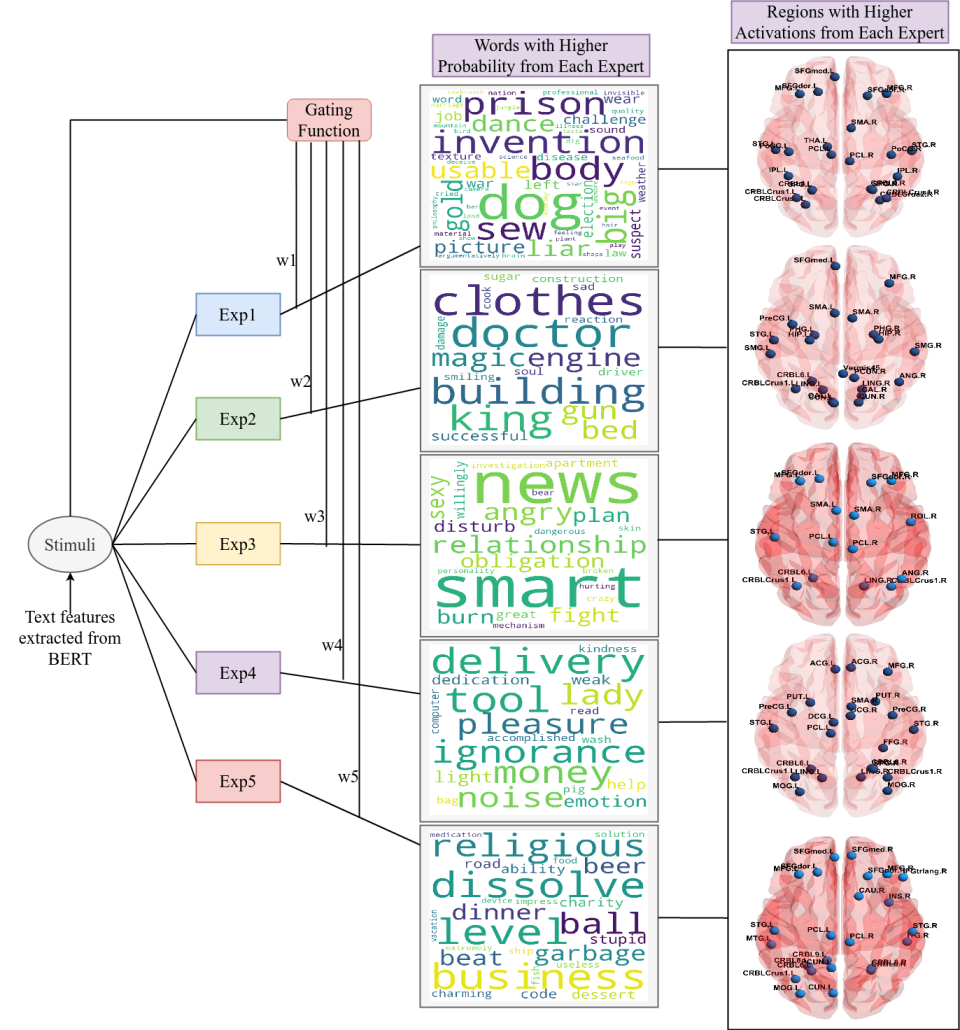}
    \caption{MoRE Architecture}
    \label{fig:architecture}
\end{figure*}

We use a mixture of experts-based encoder model whose architecture is inspired from~\cite{jordan1995convergence}. The mixture of experts architecture is composed of a gating network and several expert networks, each of which solves a function approximation problem over a local region of the input space. 
Figure~\ref{fig:architecture} shows an overview of our model where the input is the text vector extracted from the popular pre-trained neural word embedding model, namely BERT~\cite{devlin2018bert}. 
The input feature representations are given to both the expert networks and the gating network. The gating network uses a probabilistic model to choose the best expert for a given input vector. The corresponding brain activation (for all the voxels) is used as a target vector during training.
%During testing, only text vector is given to initiate the gating network to choose the expert which yields higher probability for the corresponding stimuli. 
As a result of training, the model learns to select appropriate expert via gating parameters in order to predict the whole brain activation for a particular stimulus. Also, the model highlights the specific activated brain regions for a particular stimulus.
A similar architecture is used to build distinct models for different subjects.
In the experiments and results section, we provide an in-depth analysis of the model hyper-parameters and training.

\subsection{Architecture: }

Let $S=\{(\xx_1,\yy_1),\ldots,(\xx_N,\yy_N)\}$ denote the training set where $N$ is the number of examples. $\xx_i \in \R^{n},\;\forall i\in [N]$\footnote{$[N]$ represent the set $\{1,2,\ldots,N\}$.} are the input vectors (word embeddings). $\yy_i \in \R^{m},\;\forall i\in [N]$ are the target vectors (whole brain activation all the voxels in the fMRI images). Let $K$ be the number of experts. 
The mixtures of experts model formulates the conditional density of $\yy$ given $\xx$ as a mixture of $K$ different densities as follows.
\begin{equation}
p(\yy|\xx) = \sum_{j=1}^{K}P(j|\xx,\theta_0)p(\yy|\xx,\theta_{j}) = \sum_{j=1}^{K}g_{j}(\xx,\theta_{0})p(\yy|\xx,\theta_{j})
\end{equation}
Here, $P(j|\xx,\theta_0)=g_j(\xx,\theta_0)$ is the probability of choosing $j^{th}$ expert for a given $\xx$. Note that $\sum_{j=1}^K g_j(\xx,\theta_0) =1$ and $g_j(\xx,\theta_0)\geq 0,\;\forall j\in [K]$. $g_j(\xx,\theta_0)$ is also called the \emph{gating function} and is parameterized by $\theta_0$. $p(\yy|\xx,\theta_{j})$ denotes the density function for the output vector associated with the $j^{th}$ expert
and $\theta_j$ denotes the parameters associated with the $j^{th}$ expert. 

In this paper, we choose $p(\yy|\xx,\theta_{j})$ as multivariate Gaussian probability density for each of the experts, denoted by:
\begin{equation}
\resizebox{.91\linewidth}{!}{$
    \displaystyle
p(\yy|\xx,W_j,\Sigma_j) = \frac{1}{(2\pi)^{m/2}|\Sigma_j|^{1/2}}\exp\left(-\frac{1}{2}(\yy-W_j\xx)^{T}\Sigma_j^{-1}(\yy-W_j\xx)\right)
$}
\end{equation}

where $W_{j} \in \mathbb{R}^{m \times n}$ is the weight matrix and $\Sigma_{j} \in \mathbb{R}^{m\times m}$ is the variance-covariance matrix associated with the $j^{th}$ expert. Thus, $\theta_j=\{W_j,\Sigma_j\}$. 
In this formulation we assume that the covariance matrix $\Sigma_{j}$ is diagonal. 
Thus, $\Sigma_j = \text{diag}(\sigma_{j,1}^2,\sigma_{j,2}^2,\ldots,\sigma_{j,m}^2)$, $\forall j\in [K]$.
Thus, we assume that the components of the output vector ${\bf y} \in \mathbb{R}^m$ are statistically independent of each another. 
We use this assumption to make the model simple by reducing the number of overall parameters. This assumption also makes the algorithm computationally less expensive.
Assuming $\Sigma_j = \text{diag}(\sigma_{j,1}^2,\sigma_{j,2}^2,\ldots,\sigma_{j,m}^2)$, we rewrite the conditional probability density model for $j^{th}$ expert as follows:
\begin{align*}
P({\bf y}|{\bf x},W_j,\Sigma_{j}) = \frac{1}{(2\pi)^{m/2}\sigma_{j,1}\sigma_{j,2}..\sigma_{j,m}}\exp\left(-\sum_{i=1}^m \frac{(y_i-{\bf w}_{j,i}^T\xx)^2}{2\sigma_{j,i}^2} \right)
\end{align*}
where ${\bf w}_{j,i}$ is the $i^{th}$ row of $W_j$. We use softmax function for the gating variable $g_j(\xx,\theta_0)$.
\begin{equation*}
g_{j}(\xx,\theta_{0}) = \frac{\exp\left({\bf v}_j^T\xx\right)}{\sum_{i=1}^{K}\exp\left({\bf v}_i^T\xx\right)}
\end{equation*}
where ${\bf v}_j \in \mathbb{R}^n,\;\forall j \in [K]$. Thus, $\theta_0=\{{\bf v}_1,\ldots,{\bf v}_K\}$.
Let $\Theta$ be the set of all the parameters involved for the K-experts. Thus, $\Theta=\{\theta_{0},(W_{1},\Sigma_{1}),\ldots,(W_K,\Sigma_K)\}$.

%\textcolor{blue}{\bf Finished}
%\textcolor{red}{{\bf Although we assume $\Sigma$ to be diagonal, when we construct and evaluate various models we consider different configurations in Section~\ref{sec: BIC_based Choosing}, such as \emph{full}, \emph{tied}, \emph{diag}, and \emph{spherical}. These configurations correspond to different kinds of assumptions on the elements of the covariance matrices of the experts.CORRECT PL CHECK??}}

\subsection{Training Mixture of Experts Using Expectation Maximization (EM) Algorithm}
The EM algorithm is an iterative method for finding the maximum likelihood estimate (MLE) of the parameters of a probability model.

\subsection*{E-Step}  
In the E-step, we find the expectation of the complete log-likelihood.

\iffalse
\begin{align*}
Q(\Theta|\Theta^{(p)}) = \sum_{n=1}^{N}\sum_{j=1}^{K}h_{j}^{(p)}(\xx_n) \left[\log(g_{j}(\xx_{n},\theta_{0}))+ 
\log(P(\yy_{n}|\xx_{n},W_{j},\Sigma_{j}))\right]
\label{eq:QFunc}
\end{align*}
\fi

\begin{align}
  & Q(\Theta|\Theta^{(p)})
  =
  \!\begin{aligned}[t]\medmath{\sum_{n=1}^{N}\sum_{j=1}^{K}h_{j}^{(p)}(\xx_n)} & \medmath{[\log(g_{j}(\xx_{n},\theta_{0}))}\\
  & +\medmath{\log(P(\yy_{n}|\xx_{n},W_{j},\Sigma_{j}))]} 
  \label{eq:QFunc}
  \end{aligned} 
\end{align}

where $p$ is the iteration index and $h_{j}^{(p)}(\xx_n)$ is given by
\begin{align*}
h_{j}^{(p)}(\xx_n) 	= \frac{g_{j}(\xx_n,\theta_{0}^{(p)})P(\yy_n|\xx_n,W_{j}^{(p)},\Sigma_{j}^{(p)})}{\sum_{i=1}^{K}g_{i}(\xx_n,\theta_{0}^{(p)})P(\yy_n|\xx_n,W_{i}^{(p)},\Sigma_{i}^{(p)})} 
\end{align*}

\subsection*{M-Step}
The M step chooses a parameter $\Theta$ that maximizes $Q$ function (given in eq.(\ref{eq:QFunc})). Thus,
\begin{align*}
    \Theta^{(p+1)}=\underset{\Theta}{\text{argmax}} \hspace{2pt} Q(\Theta|\Theta^{(p)})
\end{align*}

\begin{enumerate}
\item {\bf Updating $\theta_0$: }We use gradient ascent to maximize $Q$ function with respect to parameters $\theta_0$ as there does not exist any closed-form solution for the maximizer.
\begin{align*}
{\bf v}_{j}^{(p+1)} &= {\bf v}_{j}^{(p)}+\eta \nabla_{{\bf v}_j}Q(\Theta|\Theta^{(p)})\\
&= {\bf v}_{j}^{(p)}+\eta\sum_{n=1}^{N}[h_{j}^{(p)}(\xx_n)-g_{j}(\xx_n,\theta_{0}^{(p)})]\xx_{n}
\end{align*}
where $\eta$ is the step size.
\item {\bf Updating $W_{j}$: }$W_j$  comprises $m$ rows $\ww_{j,i}$. We derived the closed-form solution for $\ww_{j,i}^{(p+1)}$ as follows:

\begin{align*}
& \ww_{j,i}^{(p+1)} 
= 
\!\begin{aligned}[t]\medmath{\left[\sum_{n=1}^N h^{(p)}_j(\xx_n)\;\xx_n \xx_n^T\right]^{-1}} & \medmath{\left[\sum_{n=1}^N h^{(p)}_j(\xx_n)\;y_{n,i}\xx_n\right]}; \\
& \medmath{\;j\in[K];\;i\in [m]}
\end{aligned}
%\eta\nabla_{W_j}Q(\Theta|\Theta^{(l)})\\
%&= W_{j}^{(p)}+\eta \sum_{n=1}^{N}\sum_{j=1}^{K}h_{j}^{(p)}(\xx_n) \xx_n\Sigma_j^{-1}(\yy^{(n)}-W_{j}\xx_n)
\end{align*}

where $y_{n,i}$ is the $i^{th}$  element of ${\bf y}_n$.
\item {\bf Updating $\Sigma_j$: }
$\Sigma_j$ comprises  $\sigma_{j,1},\ldots,\sigma_{j,m}$. We derived the closed-form update equation for each of them as follows:
\begin{align*}
& \sigma_{j,i}^{(p+1)} 
= 
\!\begin{aligned}[t]\medmath{\frac{1}{\sum_{n=1}^{N}h_{j}^{(p)}(\xx_n)}\sum_{n=1}^{N}h_{j}^{(p)}(\xx_n)}&\medmath{(y_{n,i}-\ww_{j,i}^{(p)}.\xx_n)^{2}};\\
& \medmath{\;j\in[K];\;i\in [m]}
\end{aligned}
\end{align*}
\end{enumerate}

 An iteration of EM increases the original log-likelihood ${\cal L}(\Theta|\yy_1,\ldots,\yy_N)$. That is, ${\cal L}(\Theta^{(p+1)}|\yy_1,\ldots,\yy_N)>{\cal L}(\Theta^{(p)}|\yy_1,\ldots,\yy_N)$.
The likelihood ${\cal L}$ increases monotonically along the sequence of parameter estimates generated by the EM algorithm \cite{Bishop2006}.

\subsection{Selection of Number of Experts}

To find the number of experts, we used two methods.

\noindent{\textbf{Bayesian Information Criterion}} BIC  is one of the successful measures to approximate the Bayes factor~\cite{kass1995bayes}, i.e., to find a model that has maximum posterior probability or maximum marginal likelihood as well as a minimum number of model parameters. 
BIC can be formulated as follows.
\begin{align*}
   \label{bic}
    BIC = d\log(N) - 2\log({\cal L}(\Theta|\yy_1,\ldots,\yy_N))
\end{align*}
Where $d$ is the number of parameters, $N$ is the number of data points. In our proposed model, where we assume diagonal covariance matrices for each expert, $d=K\left(mn+m+n\right)$, where $K$ is the number of experts, $n$ is the dimension of input feature $\mathbf{x}$, $m$ is the dimension of the output vector $\mathbf{y}$. %\textcolor{blue}{\bf finished} %\textcolor{red}{{\bf Better to remind what $K$ (no of experts), $n$ (?) and $m$ (?) are.}} 
The objective is to find a model configuration that minimizes BIC. The model complexity increases with the increase in the number of parameters. However, the likelihood will also increase by increasing complexity. Thus, BIC makes a trade-off between the negative likelihood and the number of parameters. There exists an optimal choice of complexity (number of experts here) at which BIC takes minimum value.

\noindent{\textbf{Cross-Validation}} Since cross-validation technique was successful in predictive modeling framework~\cite{arlot2010survey}, we tested our MoRE model empirically with different number of experts -- models with 2-, 3-, 4-, and 5-experts were evaluated. 
MoRE model exhibits better separation of word categories with 5 experts compared to the other configurations. While we report the results of the model with 5-experts in the main text, those of the others are included in  supplementary Section~\ref{supplementary}.

\section{Dataset Description \& Analysis}
\label{sec:datasets}

\begin{figure*}[t]
    \begin{center}
        \includegraphics[width=1\linewidth]{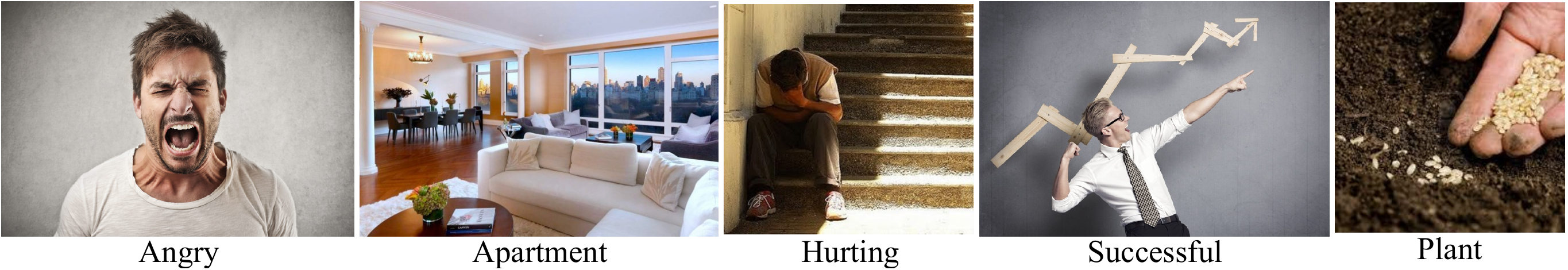}
    \caption{Examples of stimuli used for sample words (nouns, verb, and adjectives) in experiment 1. Each word along with an image is presented in multiple repetitions. These images are taken from~\cite{pereira2018toward}}
    \label{fig:dataset}
    \end{center}
\end{figure*}
In this section, we describe the dataset used for training and testing. In the next subsection, we present the process of selecting an appropriate number of experts in a model.

We used data from paradigm 1 of fMRI experiment 1~\cite{pereira2018toward}, where authors conducted experiments on multiple subjects by asking them to passively view different category of words as stimuli (images adapted from\footnote{\url{https://osf.io/crwz7}}) as shown in Figure~\ref{fig:dataset}. 
Each category might correspond to activation of distinct brain regions. 
In paradigm 1, a target word is presented along with a picture that depicted some aspect(s) of the relevant meaning. This fMRI dataset was collected from a total of 16 participants. 
%out of which 7 participants are having all three experiments data available. 
For each participant in the experiment, a whole set of 180 words (128 nouns, 22 verbs, 29 adjectives and adverbs, and one function word) were presented as stimuli, both the word along with a semantically related picture.
While participants viewed the stimuli, fMRI scans were collected. Each brain volume (scan) consists of 85 slices, each of size $88 \times 128$ voxels. 
Each participant saw stimuli, each repeated between 4 and 6 times. Here we compute the average brain response per stimulus by combining scans from all these repetitions.
%\textcolor{blue}{\bf Finished}
%\textcolor{red}{{\bf How many such scans (volumes) are collected per subject per stimulus and what is the grand total and are doing some averaging if multiple scans are collected per stimulus? We need to clarify how much data we are using in our experiments here.}}
%voxel windows [{\bf NOT CLEAR, IS IT THE SIZE OF EACH IMAGE SLICE??}] 
%% arranged as 85 slices, per subject per stimulus.
When we describe results, we depict the results of slices 10 through 77 as there is no activation in the remaining slices (the first 9 and the last 8 slices) in a brain volume. 
%% We showcase the slice results starting from $10^{th}$ slice up to $77^{th}$ slice in our results section.
%% This is due to no activations exists in the first 9 slices and last 8 slices of every brain volume.
\paragraph{Model Training: Input Representation}
We use the publicly available pre-trained word embedding method BERT~\cite{devlin2018bert} to formulate feature vectors corresponding to the 180 words used as input stimuli. Thus the BERT-based embedding yields 786-dimensional vector for each word.
\paragraph{Model Training: Output Representation}
The fMRI brain responses (voxel activation) corresponding to every stimulus were used as target output for the subject-specific encoding model. 
The dataset contains average activation (of the 201,011 voxels) for each of the 180 word-stimuli per participant.
We removed the zero intensity values from the voxels resulting in 199,658 voxels as our target output.
We follow the 5-fold cross-validation approach, where 80\% of the data (144 words) were used for training and 20\% of the data (36 words) for testing.
%\textcolor{blue}{\bf Finished}
%\textcolor{red}{{\bf Are average activations available for each of the 180 word-stimuli per participant. How much was used for training, how much held-out??}}
%%  of each participant as output to train the subject-specific model.
\paragraph{Atlas for ROI Representation}
We use the Automated Anatomical Labeling (AAL) atlas~\cite{craddock2012whole} with a parcellation of 116 brain regions to represent the brain activation response for each stimulus, where each voxel coordinate belongs to a particular region of interest.

\subsection{Choosing the Number of Experts: }
\label{sec: BIC_based Choosing}

\begin{figure}[t]
    \begin{center}
        \includegraphics[width=1\linewidth]{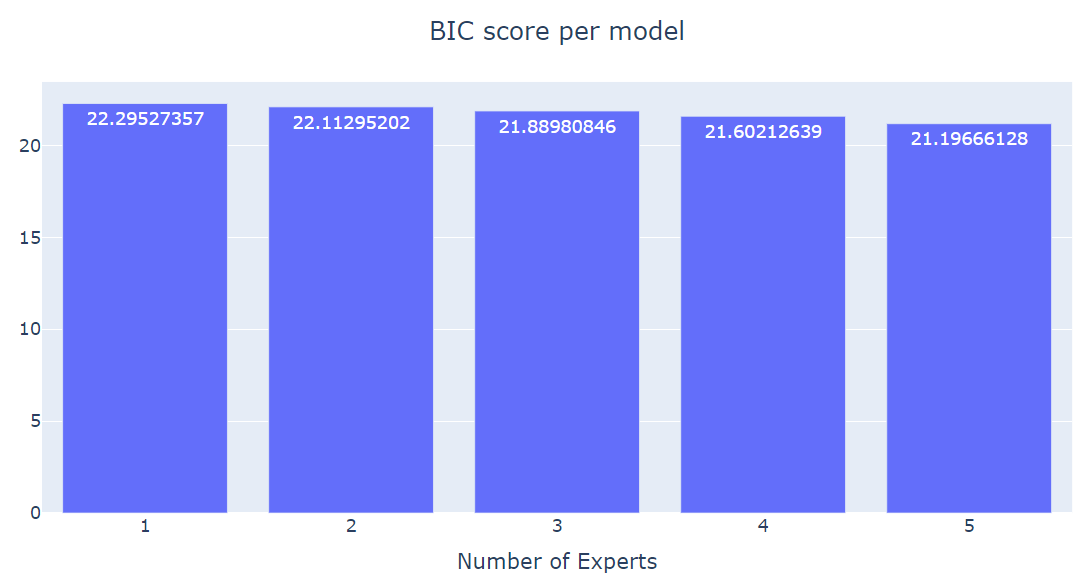}
    \caption{Selection of Number of Experts using Bayesian Information Criterion (BIC)}
    \label{fig:loglikelihood}
    \end{center}
\end{figure}

\label{sec:experts}
To get the optimal number of experts, along with cross-validation approach, we calculated the BIC scores using the diagonal covariance matrix where we consider the diagonal covariance matrix for each component. 

Here, we consider the logarithm of BIC values in Figure~\ref{fig:loglikelihood} for better visualization.
Although, the BIC values in Figure~\ref{fig:loglikelihood} seem to be  approximately similar, the model with 5 experts has a lower BIC value of 21.19.
%\textcolor{blue}{\bf Finished}
%\textcolor{red}{{\bf BIC is to be minimized but looks like you are arguing 5-expert model is better because BIC score is higher. Is BIC score defined differently from what is shown in BIC equation. Please make this clear.}}
We observe a similar pattern across all the 16 subjects, that is, the optimal model is found to be the one with 5 experts. 

\section{Data Exploratory Experiments}
\label{headings}
Before embarking on the training of encoding models, we wanted to investigate the regularities inherent in the original stimulus representations using BERT-embedding as well as the representational similarity of brain responses corresponding to different stimuli. In this section, we describe two exploratory experiments to investigate the nature of the semantic relatedness among the stimulus data as well as among the related brain response data.   
   
\subsection{Semantic Relatedness of the Stimulus Word Vectors}
\label{word_clusters}

\begin{figure}[!htb]
    \begin{center}
        \includegraphics[width=1\linewidth]{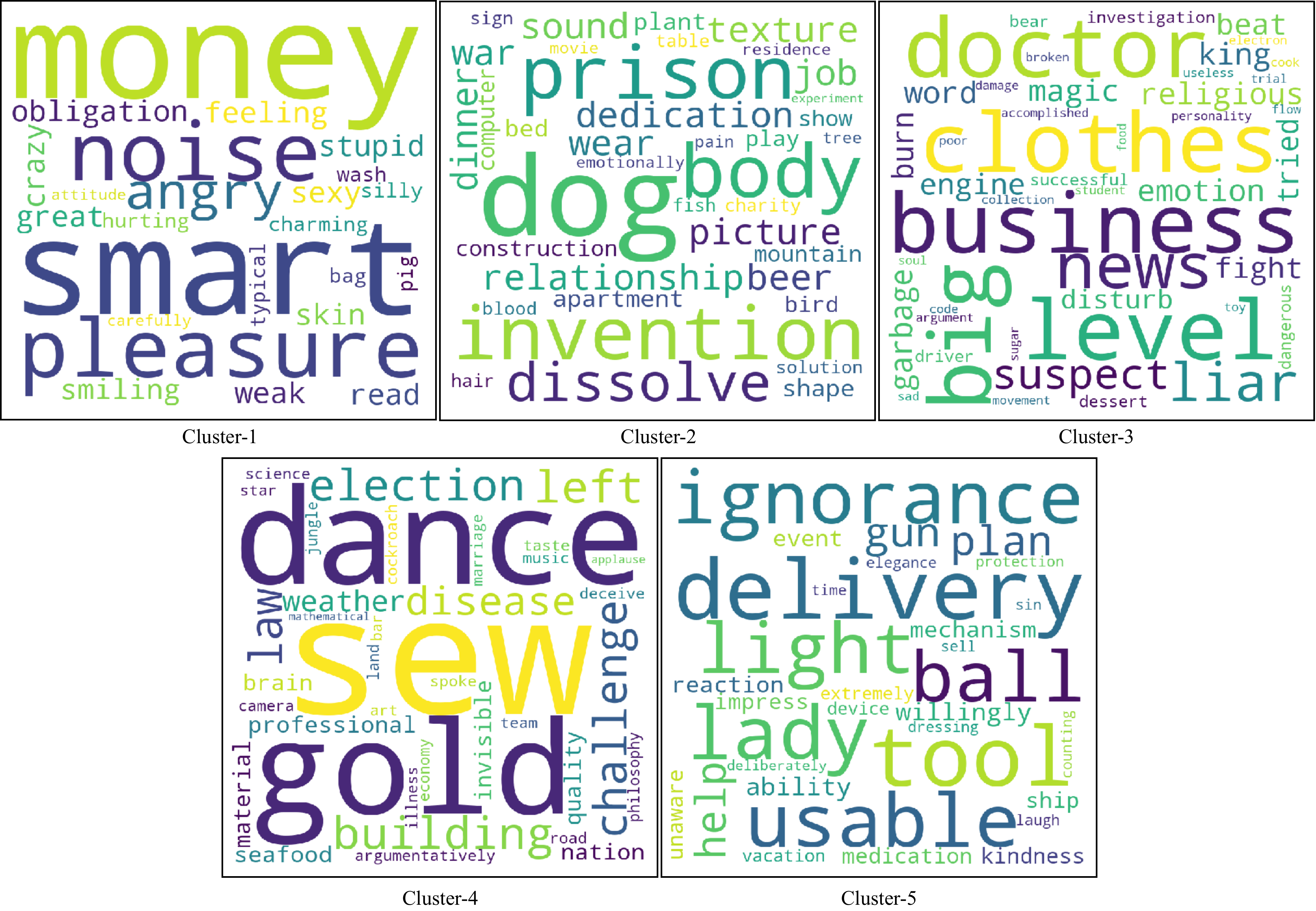}
    \caption{Visualization of the 180 target words (BERT embeddings) grouped into 5 clusters based on Cosine Similarity.}
    \label{fig:bert_words}
    \end{center}
\end{figure}

We characterize the semantic relatedness of the words used as stimuli by performing clustering of 180-word stimuli (features extracted from BERT). As seen in in the word-cloud visualization in Figure~\ref{fig:bert_words}, related words tend to appear together.
Here, we choose five clusters based on observation that the optimal model had 5 experts as described in Section~\ref{sec:experts}. We use K-means algorithm for clustering.
%\textcolor{blue}{\bf Finished}
%\textcolor{red}{{\bf What clustering procedure was used here, some details needed.}}

These word clusters provide insight into how some words are highly correlated when cosine similarity (or correlation) measure was used to investigate semantic relatedness among the word-embedding vectors.
We chose the number of clusters using the BIC \& cross-validation methods described above. 
From Figure~\ref{fig:bert_words}, we observe that semantic word pairs such as (``great'',   ``smart'', ``pleasure'', ``charming'', ``hurting''  \& ``feeling") are grouped together in Cluster 1. Similarly in Cluster 4, we have (``mathematical'', ``science'', ``law'', ``professional'', ``philosophy'',  \& ``economy"), (``election" \& ``nation"), etc. Although, we often find similar pairs in the same cluster, we also find few uncorrelated words in every cluster. %\textcolor{red}{{\bf GENERAL COMMENT: THE BERT EMBEDDINGS SEEM VAGUE. I do not really see much relatedness in the clusters shown in Fig.4, agree with your latter comment!!}}

\subsection{Clustering of the fMRI Brain Activation Vectors}
\label{corr_brain_act}

\begin{figure}[!htb]
    \begin{center}
        \includegraphics[width=1\linewidth]{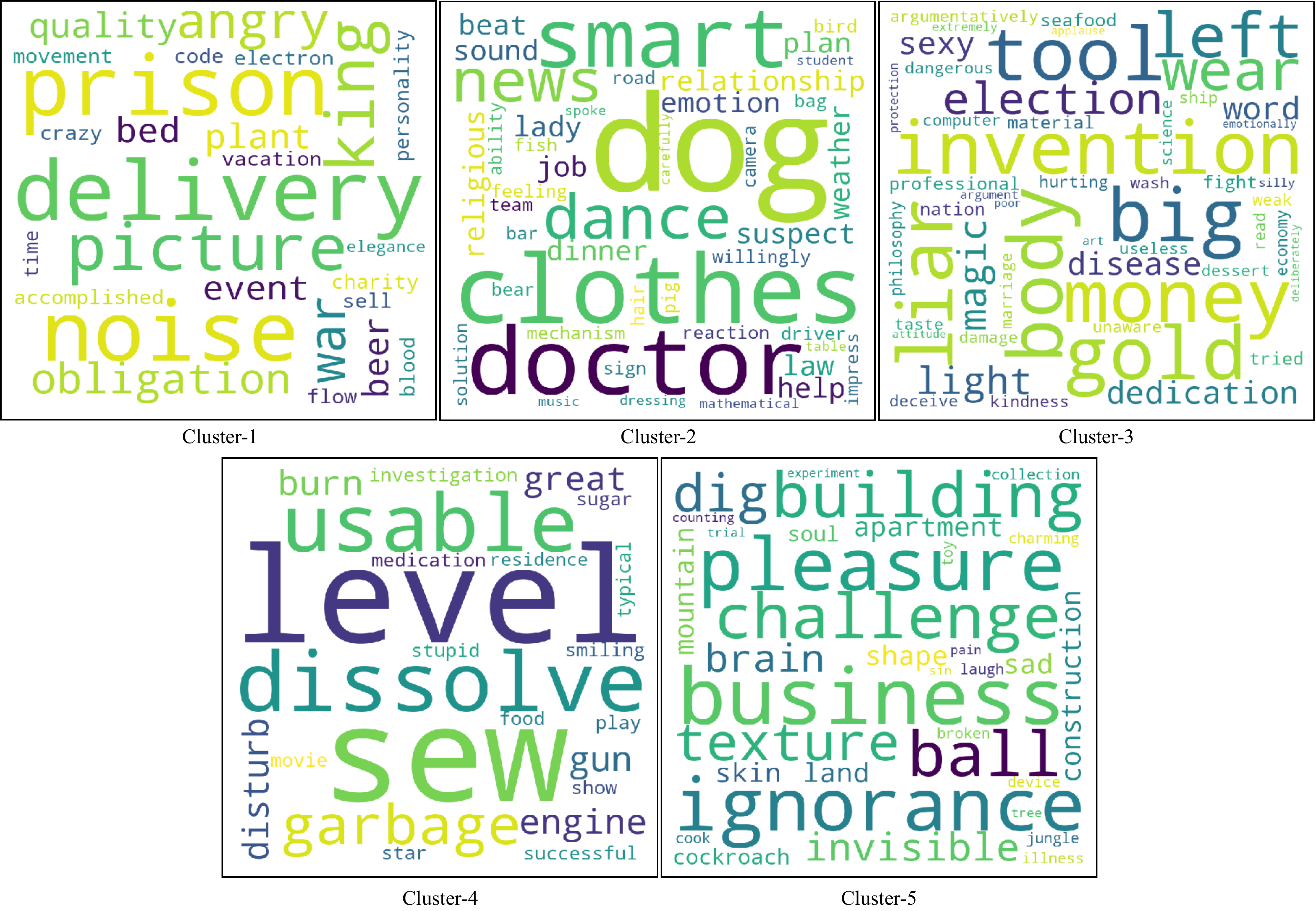}
    \caption{Visualization of the clusters formed when 180 brain activation responses corresponding to the word stimuli are grouped into 5 clusters.}
    \label{fig:voxel_words}
    \end{center}
\end{figure}

Understanding how the brain represents semantics is still in its relative infancy: how concepts are represented and combined is unclear and how this is manifested in the brain activation when subjects view the words passively is still an enigma. 
%% there are hypotheses no clearly articulated about 
In order to understand the representational relatedness of the brain activation, we performed clustering of the fMRI brain activation vectors corresponding to the 180 stimulus words. The results are shown in Figure~\ref{fig:voxel_words}. 
Similar to  section~\ref{word_clusters}, we use K-means clustering and the number clusters considered is 5, based on the observation that optimal model had 5 experts as described in section~\ref{sec:experts}.
Thus we consider the number of clusters as 5 using the BIC method as well as the cross-validation method described above. 
%\textcolor{blue}{\bf Finished}
%\textcolor{red}{{\bf See above comments on details of clustering method used.}} 
From Figure~\ref{fig:voxel_words}, we can observe that words ``skin" and ``brain" fall in Cluster 5, whereas the word ``body" falls in Cluster 3.
This result indicates that the semantic representations reflected in the brain activity might depend on the nature of the input stimuli that are shown to the subject and further might be related to how the subject responds to the image and word combination with his/her imagination and personal experience. One striking observation is that the semantic relatedness captured in Figure~\ref{fig:bert_words} is distinctly different from that in Figure~\ref{fig:voxel_words}, except for a few odd words captured in the same cluster (for example, words such as doctor, clothes, religious find themselves in the same cluster in both the cases). This is understandable as the BERT-embedding captures word co-occurrence relationship whereas brain activation is related, among other things, to the personal experience of the subject. Another possibility why these representations are different could be the ambiguity of the word-picture pairs displayed as stimuli, for example if we observe from  Figure~\ref{fig:dataset} the image is shown for the word ``Plant" looks ambiguous. The subject might be thinking of this as ``seed/s" that might in future become a plant, and hence the corresponding brain activation might include all of these inner responses evoked based on the word-picture stimulus.
The exploratory analysis suggests that although clusters may be formed based on similar semantic representations, however, each cluster may contain both correlated and uncorrelated words.
%Figure~\ref{fig:voxel_corr} shows the similarity (correlation) matrix among 180 brain activation vectors corresponding to the word stimuli, averaged across subjects. We observe high correlation again among similar words but also few unrelated words with high correlation, pointing to the variability in brain response when stimulated with lexical word and the visual stimulation by a corresponding image.

In the next section, we present the results of the proposed MoRE models as to how each expert learns the associative relationships among the word embedding and the corresponding brain activation responses.

\section{Experimental Results and Discussion}
\label{sec:results}
Here, we used paradigm-1 of experiment-1, where the target word was presented along with a related visual image (picture).
At the input level, we considered the BERT-based pre-trained embedding for extracting the features (768 dimensions) to train the models corresponding to the  three methods, namely the proposed MoRE model compared with Ridge Regression model and multilayer perceptron (MLP). The target output (brain activations) for all the methods has a very high dimension of 199658 (corresponding to the number of voxels in a brain scan).

Although we use log-likelihood in MoRE method, mean squared error (MSE) in Ridge and MLP methods as loss functions, we also consider the mean absolute error (MAE), $R^2$-score , precision, recall and F1-score as the metrics to evaluate the performance of our model. We report the the details of model performance in the next subsections.
%\textcolor{blue}{\bf Finished}
%\textcolor{red}{{\bf Where are we reporting $R^2$ scores -- this is $R^2$, not R2 CORRECT?, precision??}}. 
%We obtained a 768-dimensional vector using BERT-based pre-trained embedding and utilized this as input to all the models.  \textcolor{blue}{\bf Finished}. \textcolor{red}{{\bf What is the output dimension for all the models??}}.
We split the stimulus data into 144 words used in training and the remaining 36 words as the testing set. The encoding performance was evaluated by training and testing models using different subsets of the 180 words in a 5-fold cross-validation scheme. 
The encoder models were trained until the model reached convergence with a lower error bound of $1e^{-10}$ or till a maximum number of 200 iterations. In order to systematically present the results, the results section is divided into six parts:
%To evaluate our model performance, the experiments have been divided into six parts.
Section~\ref{baselinemethods} discusses the training performance of baseline (classical) methods: linear regression~\cite{mitchell2004learning} and MLP~\cite{oota2018fmri} for predicting the brain activity patterns.
In Section~\ref{moretraining}, we describe the training results of the proposed MoRE model for predicting the brain activations.
We present the comparison of classical results and the proposed MoRE method in Section~\ref{wordpicture}.
Section~\ref{statstical_comp} illustrates the statistical comparison between MoRE and Baseline methods. After presenting the comparative results, in Section~\ref{more_rois} we investigate the results of the MoRE model to highlight what each expert has actually learned in terms of the brain regions and associated words. Then we present the ROI predictions of the model for unseen words in Section~\ref{unknown_words}.

\subsection{Training of Baseline Methods}
\label{baselinemethods}
\begin{figure*}[!htb]
  \centering
  \includegraphics[width=0.9\linewidth]{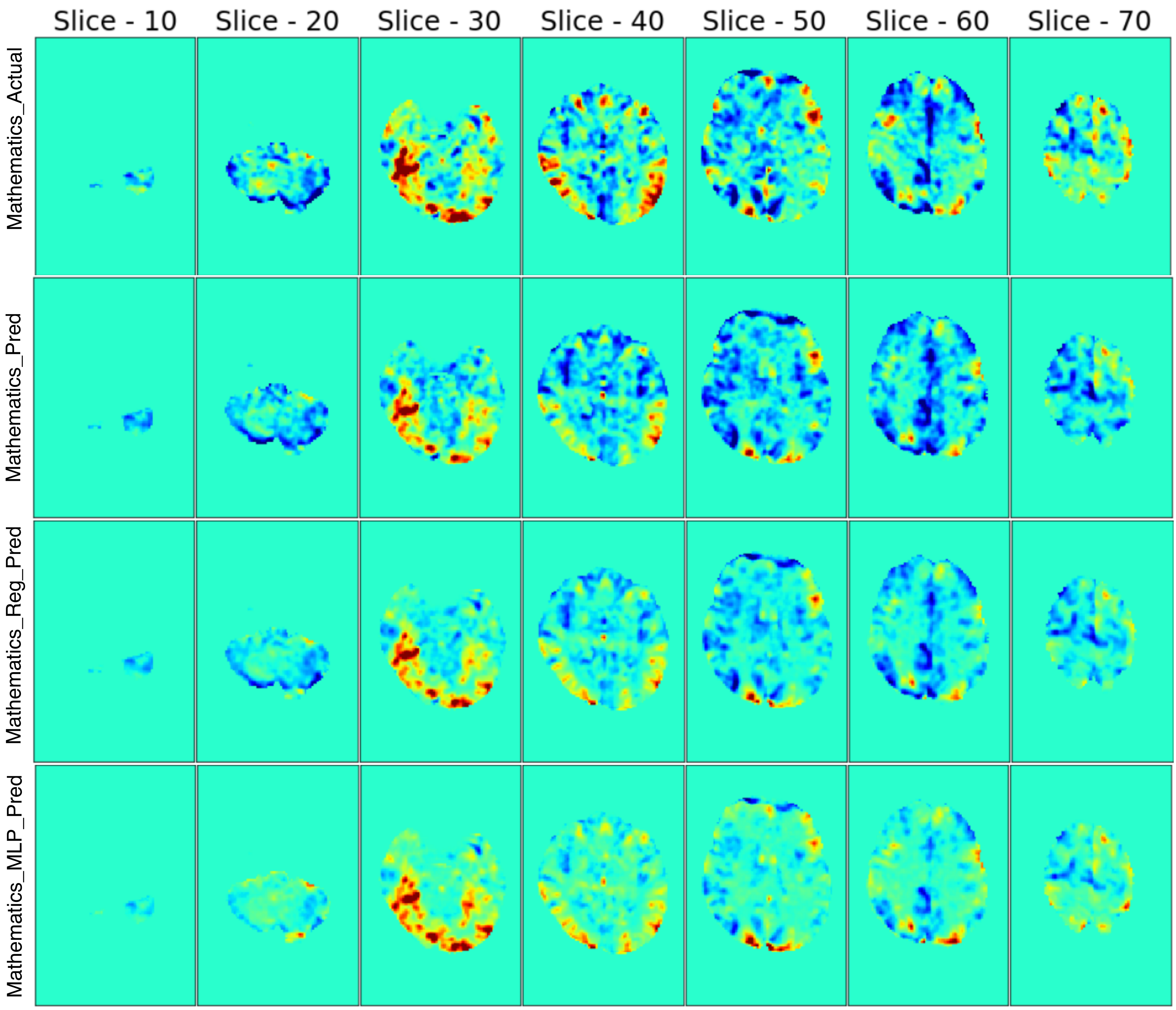}
  \caption{The figure shows variation in the fMRI slice activation actually observed and in comparison with those captured by different models for semantically related keywords.  We selected 7 random slices from the sequence of 85 slices in a brain volume to showcase the following scenarios: (i) visualization of observed voxel activation for a semantically related testing word ``mathematics'' (top row), (ii) visualization of predictions by the same expert [in the proposed 5-expert MoRE model] for semantically related testing word ``mathematics'' (second row), (iii) visualization of predicted voxels using the model trained with ridge regression (third row), and (iv) visualization of predicted voxels using the model trained with MLP (bottom row).}
  \label{fig:brainresponse}
\end{figure*}

\paragraph{Ridge Regression Method}
In the literature, linear regression has often been successfully used as a simple encoding model by the neuroscience community~\cite{mitchell2004learning,holdgraf2017encoding}. With the semantic features extracted from BERT-base as input as well as fMRI blood oxygen-level dependent (BOLD) activity collected from each subject as output, we were able to build a regression model to map the association between input (word features) and output (fMRI activations).
To avoid over-fitting, we used ridge regression method to train the model.
During model building, we used MSE as the loss function in the training, whereas mean absolute error (MAE), $R^2$-score considered as metrics to measure the model performance. 
The ridge regression method reports 5.5 MAE and 0.15 $R^2$-score.
%\textcolor{blue}{\bf Finished}
%\textcolor{red}{{\bf NOT CLEAR. WHAT ARE MAE, $R^2$ etc USED FOR DURING MODEL BUILDING, AS LOSS FNS?? IF SO WHICH ONE IS BETTER AND FINALLY ADOPTED?}}. 
Figure~\ref{fig:brainresponse} shows the predicted fMRI brain activation (fourth row) for the word ``Mathematics''.

\paragraph{Multi-layer Perceptron Method (MLP)}
In this paper, we use a 3-layer neural network to build an MLP method that predicts the neural activation pattern for a given stimulus (word vector) at the input layer.
We use 1000-hidden neurons in the hidden layer, connection weights between the input layer and hidden layer learned through an adaptation process.
The output layer provides the prediction of fMRI activations as a weighted sum of neural activations contributed by each of the hidden layer neurons.
%\textcolor{blue}{\bf Finished}
%\textcolor{red}{{\bf Confusing terminology: what does ``step'' mean? Steps mean input layer, hidden layer and output layers?? If so, needs to be changed.}}
The predicted result for the word ``Mathematics'' is shown in Figure~\ref{fig:brainresponse} (fifth row) depicts the BOLD activations across the slices.
The model predicted brain activations are measured by using MAE, $R^2$-score as the metrics.
The MLP method yields 4.63 MAE and 0.35 $R^2$-score, performs better than Ridge regression method.
%\textcolor{blue}{\bf Finished}
%\textcolor{red}{{\bf Why should we include "Science" related predicted activity in Figure 6?? It seems out of place.}}

{\renewcommand{\arraystretch}{0.3}
\begin{table*}%
\scriptsize
\centering
\caption{Comparison of average accuracy of 16 participants i) ridge regression model, ii) multi-layer perceptron and, iii) MoRE. The macro-average results display the average of performance of each individual class (class-1: voxel coordinates with value 1, class-0: voxel coordinates with value 0). In Micro-average method, we sum up the individual true positives, false positives, and false negatives of the two classes and calculate the three metrics. The last column of the table reports the performance of class-1 (voxel coordinates with value 1).}
\label{tab:heal2}
\resizebox{\textwidth}{!}{\begin{tabular}{|l|l|c c c|c c c|c c c|}
\hline
\multicolumn{2}{|c|}{$\rightarrow$}&\multicolumn{3}{c|}{\textbf{Macro Average}}&\multicolumn{3}{c|}{\textbf{Micro Average}}&\multicolumn{3}{c|}{\textbf{Class-1}}\\
\hline 
Feature set$\downarrow$&Method$\downarrow$&Precision&Recall&F1-score&Precision&Recall&F1-score&Precision&Recall&F1-score\\
\hline
\hline
& Regression&0.76&0.57&0.59&0.99&0.99&0.99&1&1&1\\

$\mu$-3$\sigma$&MLP&0.87&0.51&0.51&0.99&0.99&0.99&1&1&1\\

&MoRE&0.62&0.62&0.58&0.99&0.99&0.99&1&1&1\\ 
\hline
&Regression&0.76&0.62&0.62&0.98&0.98&0.98&0.98&0.99&0.99\\

$\mu$-2$\sigma$&MLP&0.88&0.52&0.52&0.98&0.98&0.98&0.98&1&0.99\\

&MoRE&0.63&0.64&0.61&0.97&0.97&0.97&0.99&0.99&0.99\\
\hline
&Regression&0.76&0.67&0.67&0.89&0.89&0.89&0.93&0.92&0.93\\

$\mu$-1$\sigma$&MLP&0.86&0.55&0.56&0.90&0.90&0.90&0.92&0.92&0.92\\

&MoRE&0.66&0.67&0.67&0.89&0.90&0.89&0.94&0.93&0.93\\
\hline
&Regression&0.73&0.61&0.64&0.66&0.66&0.66&0.77&0.7&0.74\\

$\mu$&MLP&0.74&0.58&0.61&0.64&0.64&0.64&0.8&0.65&0.72\\

&MoRE&0.66&0.64&0.64&0.66&0.66&0.66&0.75&0.72&0.74\\
\hline
&Regression&0.83&0.73&0.76&0.91&0.9&0.91&0.72&0.55&0.63\\

$\mu$+1$\sigma$&MLP&0.76&0.71&0.73&0.89&0.89&0.89&0.65&0.60&0.63\\

&MoRE&0.73&0.77&0.75&0.90&0.90&0.90&0.71&0.61&0.65\\
\hline
&Regression&0.77&0.76&0.76&0.97&0.97&0.97&0.7&0.68&0.69\\

$\mu$+2$\sigma$&MLP&0.8&0.74&0.76&0.96&0.96&0.96&0.66&0.69&0.68\\

&MoRE&0.76&0.77&0.76&0.97&0.97&0.97&0.72&0.69&0.70\\ 
\hline
&Regression&0.82&0.76&0.77&0.99&0.99&0.99&0.33&0.82&0.47\\

$\mu$+3$\sigma$&MLP&0.76&0.78&0.77&0.98&0.98&0.98&0.99&0.90&0.95\\

&MoRE&0.74&0.8&0.78&0.99&0.99&0.99&0.43&0.90&0.58\\
\hline
\end{tabular}}

\end{table*}
}

\begin{figure}[t]
    \begin{center}
        \includegraphics[width=0.8\linewidth]{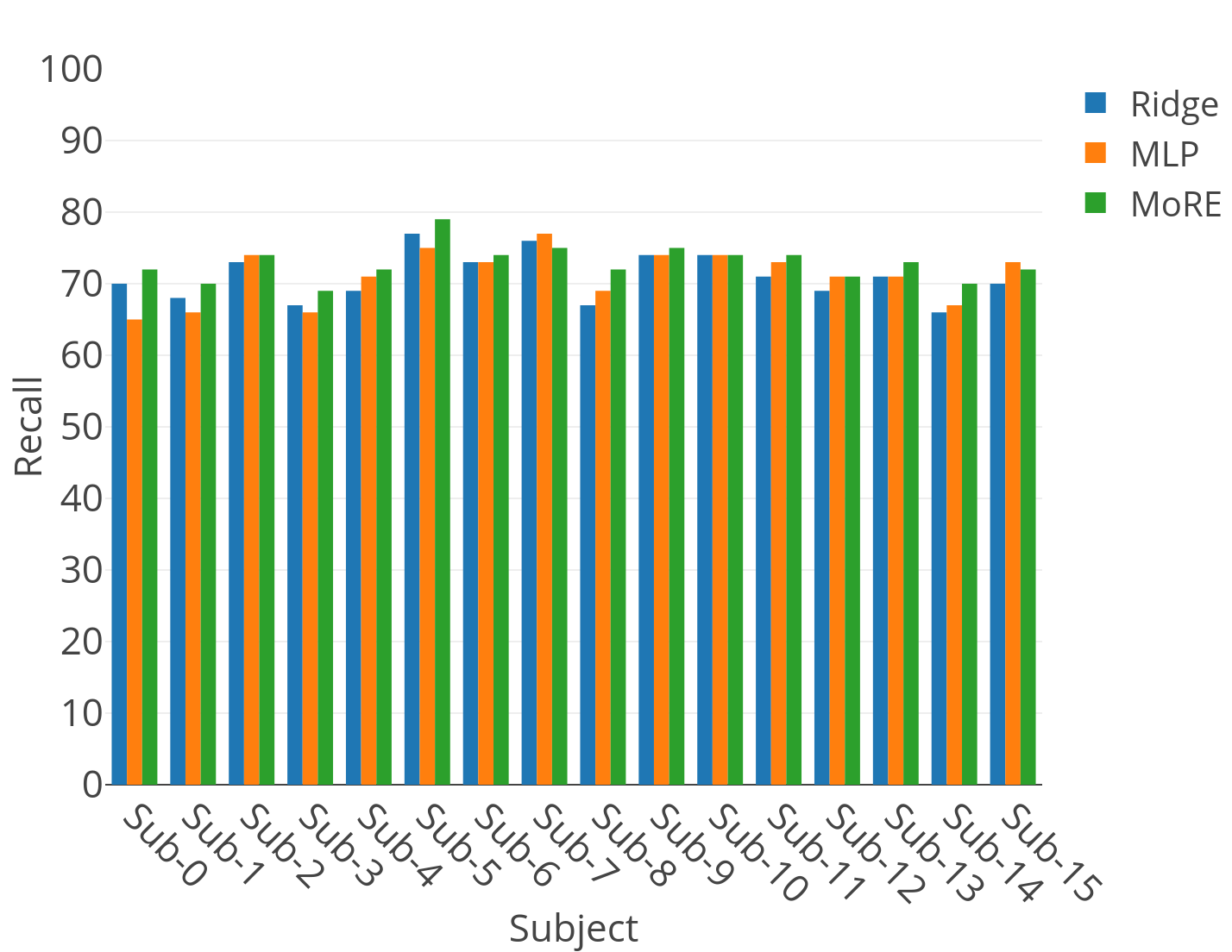}
    \caption{Comparison of recall performance of i) ridge regression model, ii) multi-layer perceptron and, iii) mixture of regression experts. The individual models built for each of the 16 participants  are shown here.}
    \label{fig:recall}
    \end{center}
\end{figure}

\begin{figure}[t]
    \begin{center}
        \includegraphics[width=0.8\linewidth]{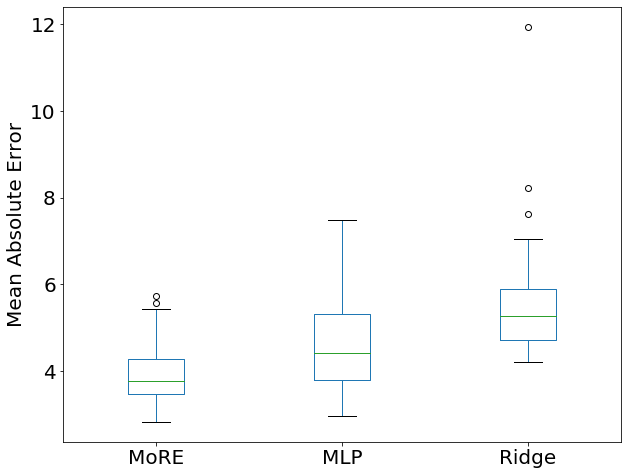}
    \caption{Box-plot for average mean absolute error of all test words for the three methods. Horizontal lines represent median ranks, and the median rank (MAE value) of MoRE method is significantly less than that of MLP and Ridge methods.  %\textcolor{blue}{{\bf finished}} \textcolor{red}{{\bf MEDIAN RANK LESS MEANS WHAT, NOT CLEAR}}
    }
    \label{fig:boxplot}
    \end{center}
\end{figure}

\begin{table*}[!htb]
  \caption{List of expert-wise keywords used during training, the keywords predicted during testing, and expert-wise regions of activation (ROIs) in the brain. 
  %The last row lists the common regions of activation among all the experts.
  } 
  
   \label{words_results}
   \centering
\resizebox{0.9\textwidth}{!}{  \begin{tabular}{|c|c|c|c|}
  \hline
  \textbf{Expert} & \textbf{Train Set} & \textbf{Test Set} & \textbf{Important Regions} \\ \hline \hline
Expert-1 &    \begin{tabular}{c} \includegraphics[width=0.27\linewidth]{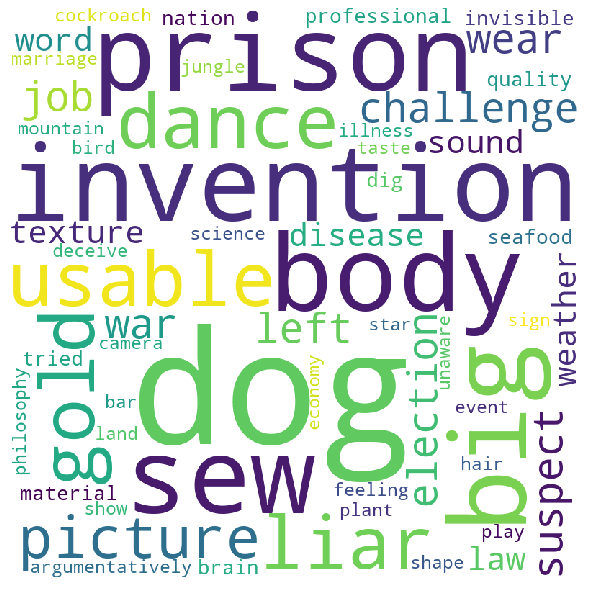} \\
\end{tabular}  & \begin{tabular}{c} \includegraphics[width=0.27\linewidth]{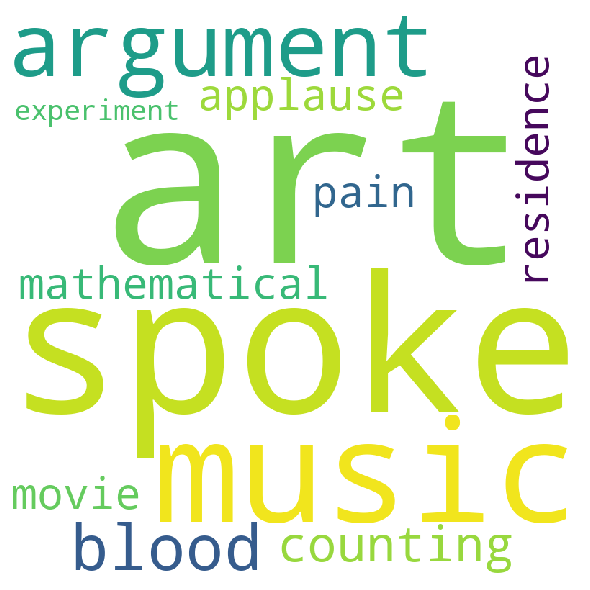} \\
\end{tabular} & \begin{tabular}{c}
\includegraphics[width=0.27\linewidth]{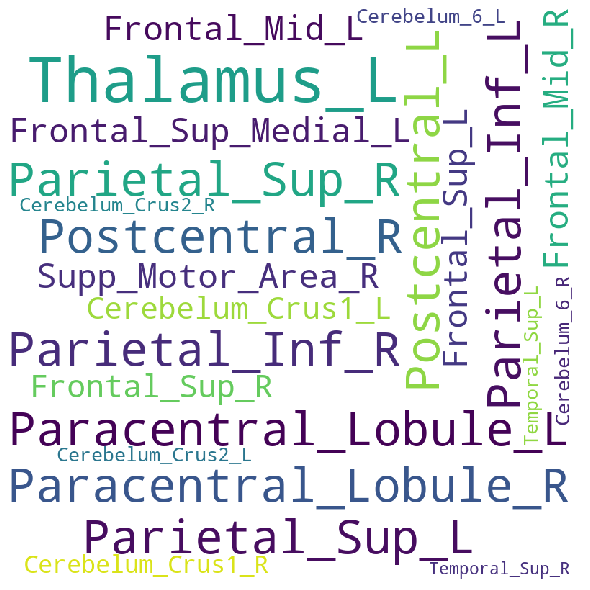} \\
\end{tabular} \\ \hline
Expert-2 &   \begin{tabular}{c}  \includegraphics[width=0.27\linewidth]{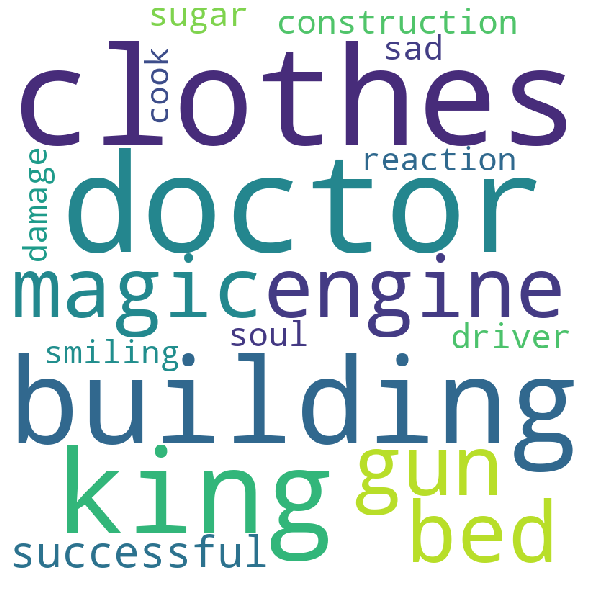} \\ \end{tabular} & \begin{tabular}{c} \includegraphics[width=0.27\linewidth]{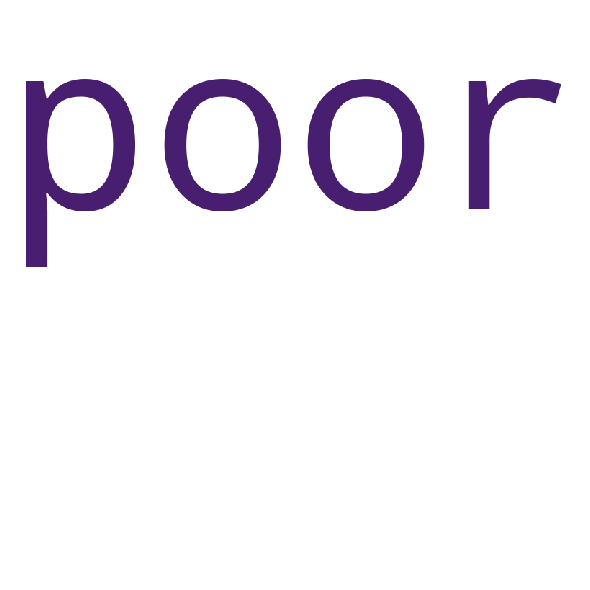} \\ \end{tabular}& \begin{tabular}{c} \includegraphics[width=0.27\linewidth]{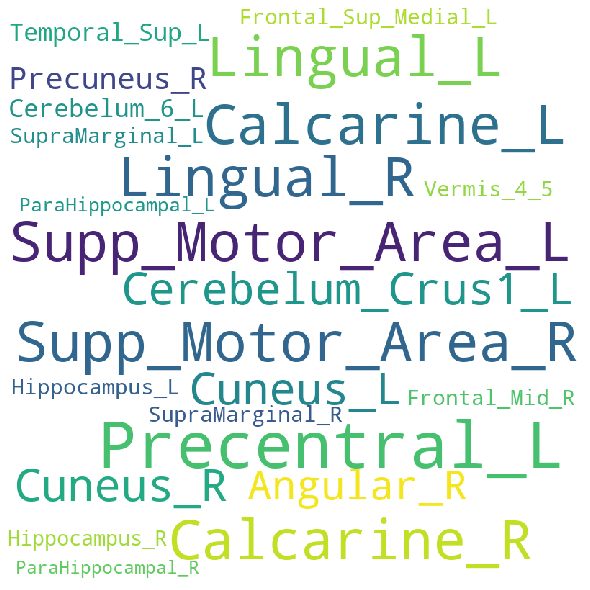} \\ \end{tabular} \\ \hline
Expert-3 &  \begin{tabular}{c}    \includegraphics[width=0.27\linewidth]{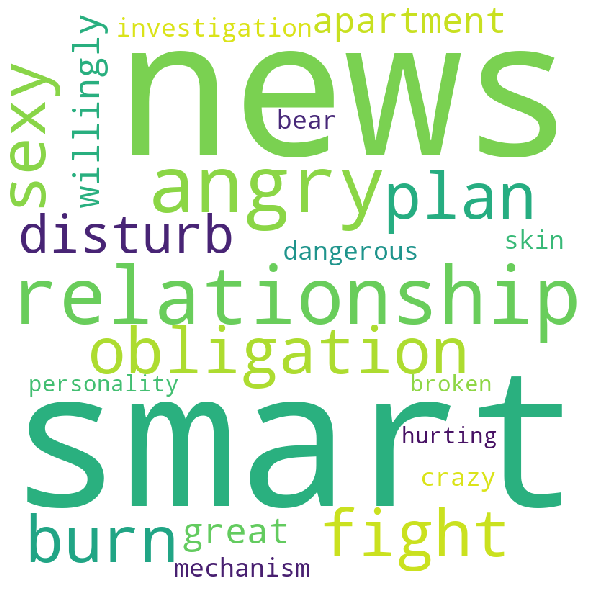} \\ \end{tabular} & \begin{tabular}{c}  \includegraphics[width=0.27\linewidth]{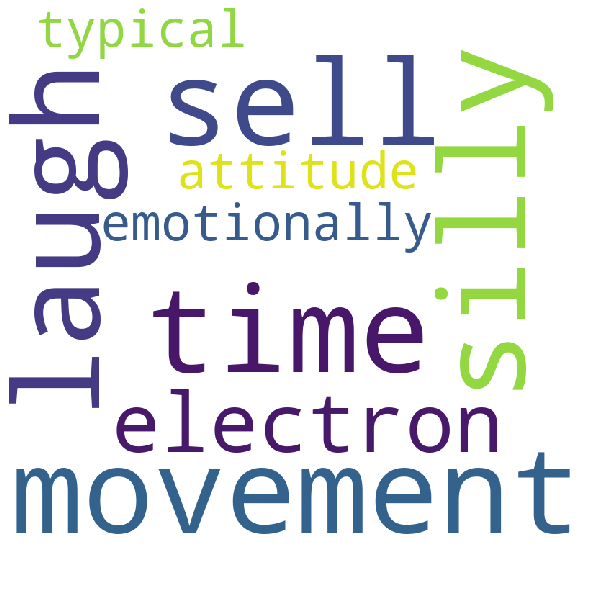} \\ \end{tabular}& \begin{tabular}{c}  \includegraphics[width=0.27\linewidth]{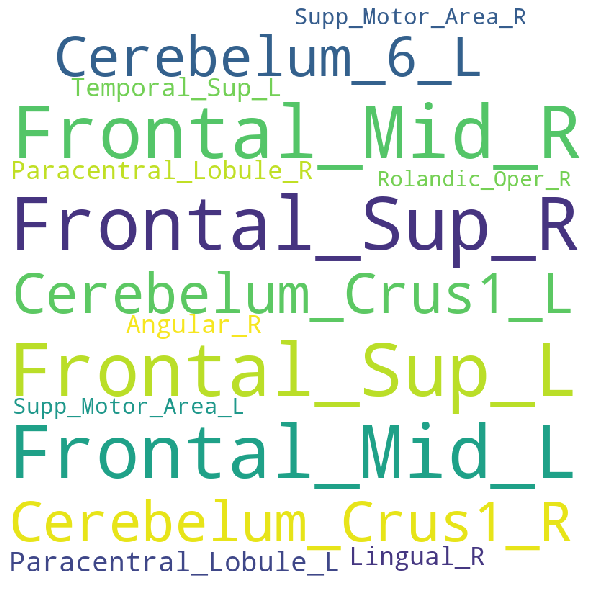} \\ \end{tabular} \\ \hline
Expert-4 &  \begin{tabular}{c}  \includegraphics[width=0.27\linewidth]{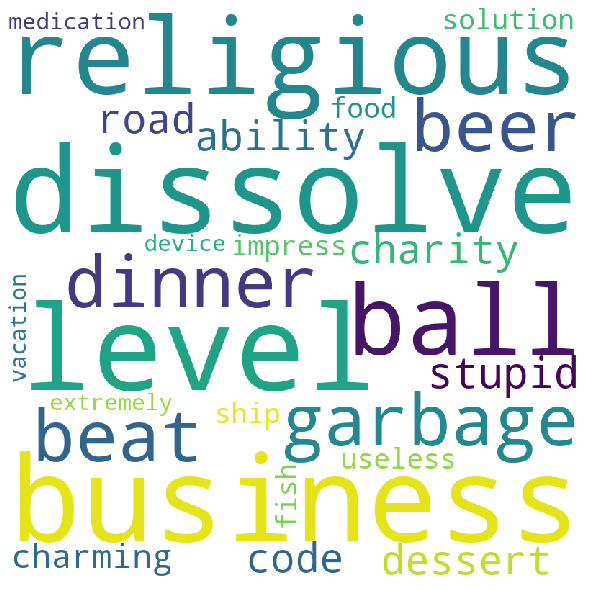} \\ \end{tabular} & \begin{tabular}{c}  \includegraphics[width=0.27\linewidth]{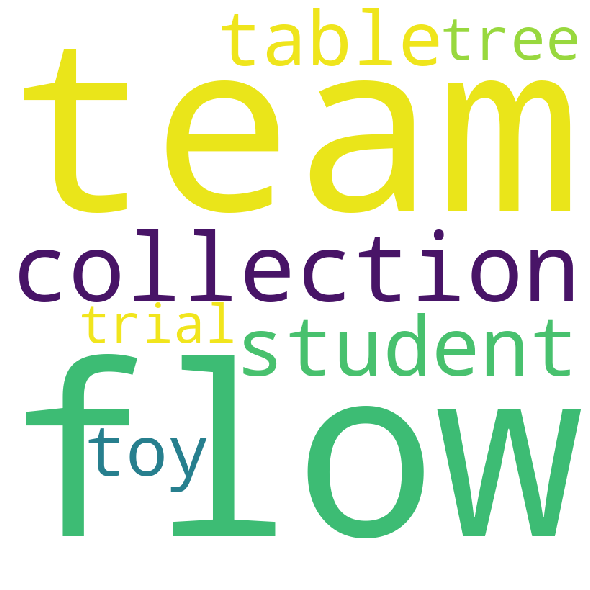} \\ \end{tabular} & \begin{tabular}{c}  \includegraphics[width=0.27\linewidth]{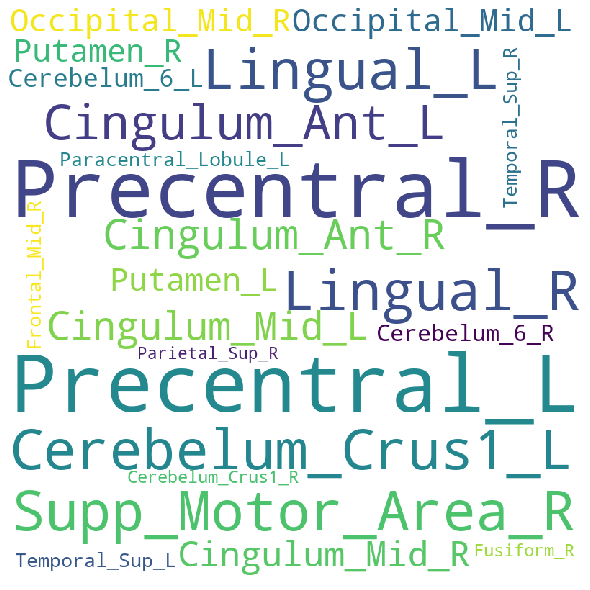} \\ \end{tabular} \\ \hline
Expert-5 &  \begin{tabular}{c}    \includegraphics[width=0.27\linewidth]{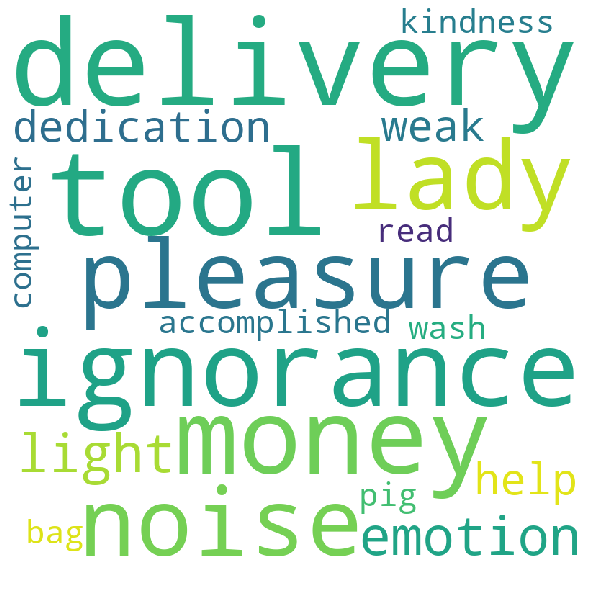} \\ \end{tabular}& \begin{tabular}{c}  \includegraphics[width=0.27\linewidth]{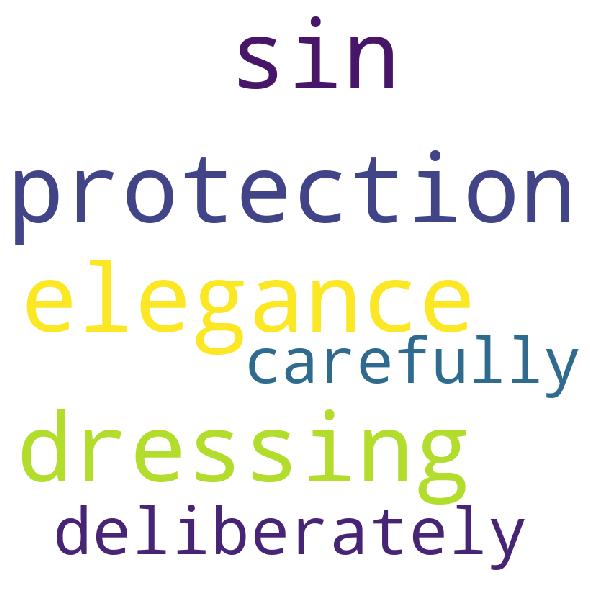} \\ \end{tabular} & \begin{tabular}{c}  \includegraphics[width=0.27\linewidth]{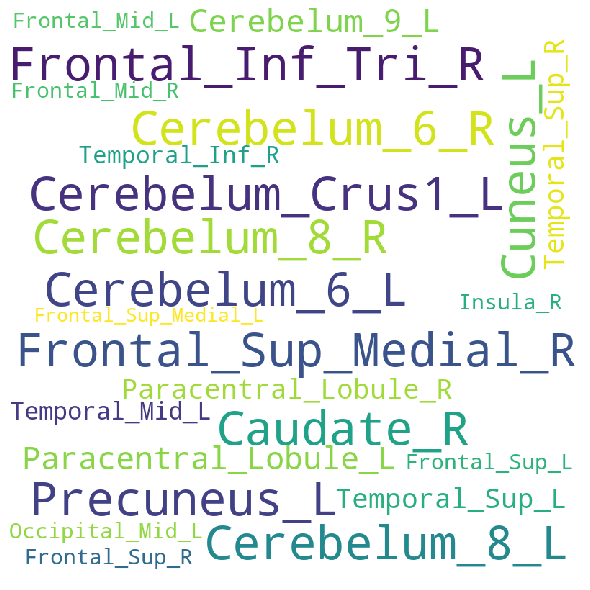} \\ \end{tabular} \\ \hline

  \end{tabular}}
\end{table*}

\subsection{MoRE Training}
\label{moretraining}

Using the approach discussed in Section~\ref{Approach} and using the insights from the experiments in Section~\ref{sec: BIC_based Choosing}, we trained a separate mixture of five-regression experts (MoRE) model for each subject.
We performed experiments on the dataset where the stimulus (text) vector extracted from the recently successful neural word-embedding method, namely, BERT, was given as input to the model and estimated the corresponding brain activation response as the output of the model.
%% Specifically, we obtained a 768-dimensional vector using BERT-base pre-trained embedding.
%We split the stimulus data into 144 words used in training and the remaining 36 words as the testing set. 
%% The encoding performance was evaluated by training and testing models using different subsets of the 180 words in a 5-fold cross-validation scheme. 
%% The encoder models were trained until the model reached convergence with a lower bound of $1e^{-10}$ or till a maximum number of 200 iterations.

%\subsection{Stimulus Presentation with Target Word and Visual Image (Picture)}
\subsection{Comparison of Results of MoRE model with those of the Baseline Models}
\label{wordpicture}
%\noindent{\textbf{Step 1: MoRE Results comparison with Baseline}} 
We predicted the brain activation using classical models as well as the proposed MoRE method.
To statistically verify how our results are reliable and finding the variance in predicted brain activations from the observed average voxel intensity values, we use three-Sigma rule implies that heuristically nearly all values lie within three standard deviations of the mean.
Table~\ref{tab:heal2} shows the macro/micro-average precision, recall, F1-score of 16 participants obtained using classical methods such as Ridge regression and MLP and compared with those of the MoRE method.
Here, the macro averaging gives equal weight to each class to evaluate the performance of various methods across the two-classes. On the other hand, micro-averaging method calculates the individual true positives, true negatives, false positives, and false negatives of the the binary-class model.
%\textcolor{blue}{\bf Finished}
%\textcolor{red}{{\bf GIVE BRIEF INTUITION HERE FOR THIS COMPLEX TESTING PROTOCOL WITH VARIOUS SIGMA's}}.
In Table~\ref{tab:heal2}, we divided the ground truth voxel intensity values of test data into the following distributions such as ``$\mu$-3$\sigma$'', ``$\mu$-2$\sigma$'', ``$\mu$-1$\sigma$'', ``$\mu$'', ``$\mu$+1$\sigma$'', ``$\mu$+2$\sigma$'', ``$\mu$+3$\sigma$'', where ``$\mu$'' is mean and ``$\sigma$'' stands for standard deviation.
For each ground-truth test word, we calculate the ``$\mu$'' and ``$\sigma$'' from the 199,658 voxel intensity values.
We make the voxel coordinates (199,658 voxels) into 2 classes for each of the above-mentioned distributions for each of the three methods.
%We repeat the same process for remaining distributions as well. 
We consider the following steps for getting the classification metrics.
%\textcolor{blue}{\bf Finished}
%\textcolor{red}{{\bf THE FOLLOWING DESCRIPTION IS VAGUE, NOT CLEAR}}
\begin{itemize}
    \item For each test word, we predicted the brain activations from the three methods such as Ridge, MLP, and MoRE.
    \item We take the predicted brain activations from the previous step and assign the intensity values higher than ``$\mu$-3$\sigma$'' to one class and the remaining voxel coordinates as belonging to the zero class.
    \item We perfrom a similar analysis on the empirically observed brain activations. We use the observed brain activations with voxel intensity values higher than ``$\mu$-3$\sigma$'' as one class and the other voxel coordinates as belonging to zero class.
    \item We compare the voxel coordinates of ground-truth and predicted from the previous two steps, and estimate precision, recall, and F1-scores for the three methods shown in Table~\ref{tab:heal2}.
    %\item For example, consider the ridge regression prediction results where fhe intensity values more than ``mean-3sigma'' as one class voxel coordinates, other voxel coordinates as zero class.
    \item We repeat the above steps for  other intensity value distributions (corresponding to the values that are 1- or 2- standard deviations above and below ``$\mu$'') for the three methods.
    %\item Further, each distribution wise, we compare the voxel coordinates of ground truth with predicted for each of the methods.
    %\item We measure the precision, recall, F1-score for each method in each distribution shown in Table~\ref{tab:heal2}.
\end{itemize}

We can observe from the Table~\ref{tab:heal2} that results with MoRE based method are better compared to classical models. 
%\textcolor{blue}{{\bf finished}}
%\textcolor{red}{{\bf CAN YOU BOLD THEM IN THE TABLE FOR CATCHING THEM EASILY. ALSO EXPLAIN MICRO, MACRO AVERAGES ETC}}. 
The macro average recall score for the MoRE method is higher than the two methods in all the seven distributions.
In contrast, precision for two baseline methods is higher than MoRE, which mainly comprises false positives (some other voxel coordinates that become activate spuriously).
Of the two baseline methods, MLP method results in more false positives as compared to Ridge regression.

We have shown the actual brain responses for the word ``mathematics'' and predicted brain activation of the three methods ($2^{nd}$ row MoRE, $3^{rd}$ row Ridge, and $4^{th}$ row MLP) for the test word ``mathematics'' in Figure~\ref{fig:brainresponse}. As shown in Figure~\ref{fig:brainresponse}, we observe that similarities between ground truth and cortical brain responses from the MoRE-based encoding model are better with a near-perfect recall, as described above.  This result encourages us to believe that MoRE  performs prediction based on a good semantic understanding of the stimulus and the associated brain activation. 

Figure~\ref{fig:recall} compares the performance of the baseline ridge regression model, MLP, and the proposed MoRE model. 
The proposed MoRE model has a better recall score for all the subjects except for subject-7 as compared to the linear and non-linear baseline models. 

\subsection{Statistical Analysis of the three methods} 
\label{statstical_comp}
To compare the three methods statistically, one-way ANOVA  was performed on MAE of the three methods. Results confirmed that the three methods were statistically significantly different, with an F-statistic [F(2,3) = 19.99, $p$ = 4.4e-10] % \textcolor{blue}{{\bf finished}}
%\textcolor{red}{{\bf PLEASE CHECK REPORTING STYLE FOR ANOVA RESULTS. It should be like F(2,??) = 27, p $<$ .0001 for example. Also actual post-hoc test results for significantly different pairs need to be mentioned. For example, MoRE vs MLP and MoRE vs RR whereas other pair RR vs MLP may not be significant???}}.
We use \emph{post-hoc} Scheffe’s test to the obtain the results between
different pairs such as MoRE vs MLP ($p$ = 0.0263), MoRE vs Ridge ($p$ = 2.16e-8), and Ridge vs MLP ($p$ = 0.0011), again reiterating the superior performance of the proposed model with respect to baseline methods.
Additionally, from Figure~\ref{fig:boxplot} we can also observe that the average MAE error for the models using MoRE, Ridge, and MLP are significantly different, with MoRE performance being superior to other methods.

%\noindent{\textbf{Step 2: MoRE ROIs Learned by Each Expert}}
\subsection{MoRE ROIs Learned by Each Expert}
\label{more_rois}

In the previous section comparative study of the proposed MoRE model with baseline methods suggested superior performance of the proposed model. In this section we delve deeper into what the model actually learns.
As we observe the similar set of pair of words for training and testing in each fold, the words illustrated in Table~\ref{words_results} corresponding to one of the 5-fold data.
Table~\ref{words_results} displays the words that are categorized by each expert in the train and test datasets and the corresponding highly activated brain regions. 
%Here, the words illustrated in Table~\ref{words_results} corresponding to one of the 5-fold data, where we present the expert-wise results of training, testing, and learned brain regions.
%\textcolor{blue}{{\bf finished}}
%\textcolor{red}{{\bf CAN YOU BEAUTIFY TABLE 2. Vertically center the Expert nos and bold font them as well as Column Headings. Reduce the horizontal lines and remove the vertical lines if possible}}.
From Table~\ref{words_results}, we observe that each expert exhibited distinct associations between word stimuli and the corresponding brain activation in both training and testing experiments. For example, while expert-1 has a higher probability for the word ``science", the same expert displays higher probability for a semantically related test word ``mathematical". 
To identify the brain regions which are associated with words in each expert, the following steps were taken.
\begin{itemize}
    \item We generated a matrix of size [\#No of high-probable words for each expert $\times$ Brain ROIs (regions corresponding to the activated voxels)] for that expert. \\
    For example, we can see from Table~\ref{words_results} that Expert-2 captured 18 words in training (with 116 corresponding ROIs), yielding a matrix of size $18 \times 116$. 
    \item To identify the ROIs associated with each expert, we applied principal component analysis (PCA) on the above matrix (\#words $\times$ \#regions) and extracted principal components (PCs) with a maximum explained variance ratio of 85\% resulting in a matrix (\#words $\times$ \#components).
    \item This matrix would enable us to identify the most critical variables in the original feature space that have a maximal contribution. 
    \item The region's importance is calculated by using the simple matrix multiplication of (\#regions $\times$ \#words) and (\#words $\times$ \#components). Note that the former is the transpose of the matrix whose PCA was performed in the earlier steps.
    \item By considering the features that have a positive magnitude, from each component we select the regions which have a score greater than 0.2.
    \item This process is repeated for each of the experts and column-4 in Table~\ref{words_results} displays the most important regions corresponding to a particular expert.
\end{itemize}

\begin{figure*}[t]
\minipage{\textwidth}
\centering
  \includegraphics[width=0.9\linewidth]{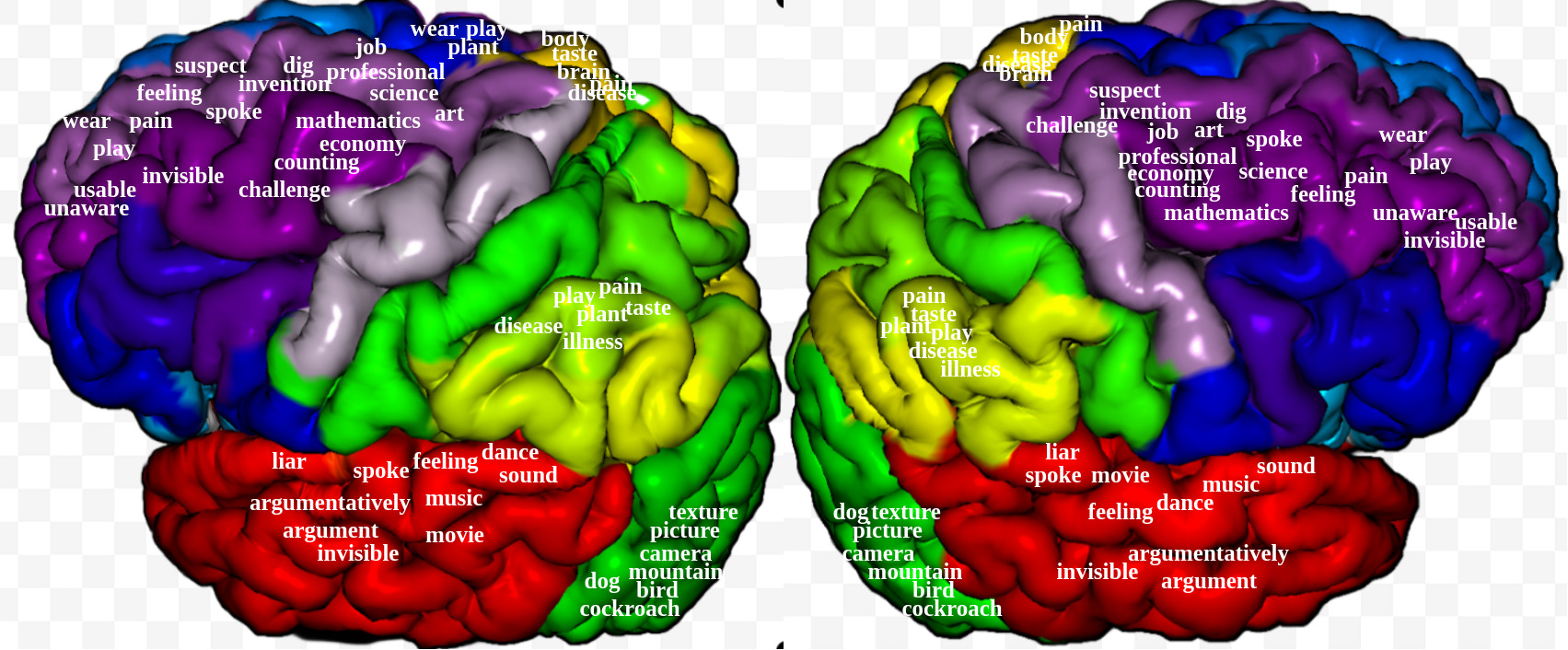}
\endminipage\hfill
\caption{Specialization of Expert-1 for words and the corresponding regions of interest (ROIs) in the brain}
\label{fig:brainroisexp1}
\end{figure*}

\begin{figure*}[t]
\minipage{\textwidth}
\centering
  \includegraphics[width=0.9\linewidth]{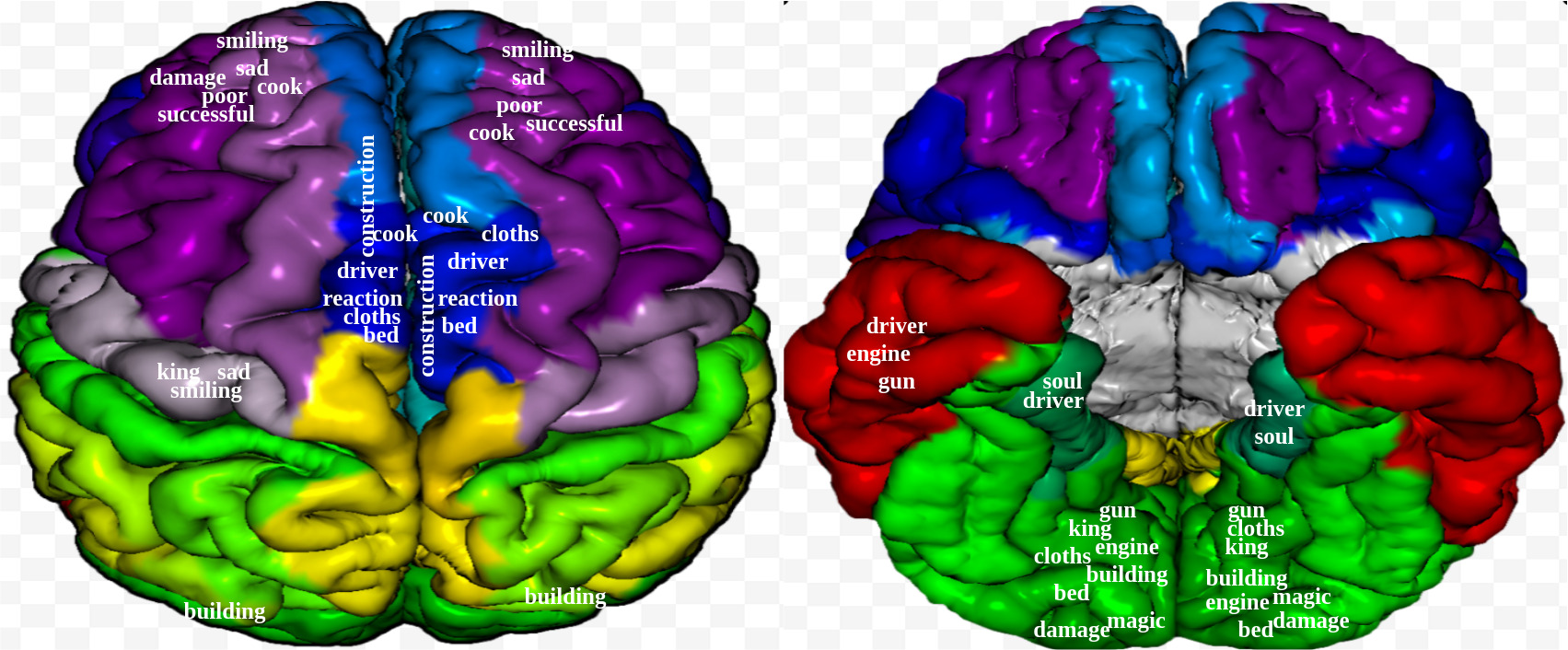}
\endminipage\hfill
\caption{Specialization of Expert-2 for words and the corresponding regions of interest (ROIs) in the brain}
\label{fig:brainroisexp2}
\end{figure*}

\begin{figure*}[t]
\minipage{\textwidth}
\centering
  \includegraphics[width=0.9\linewidth]{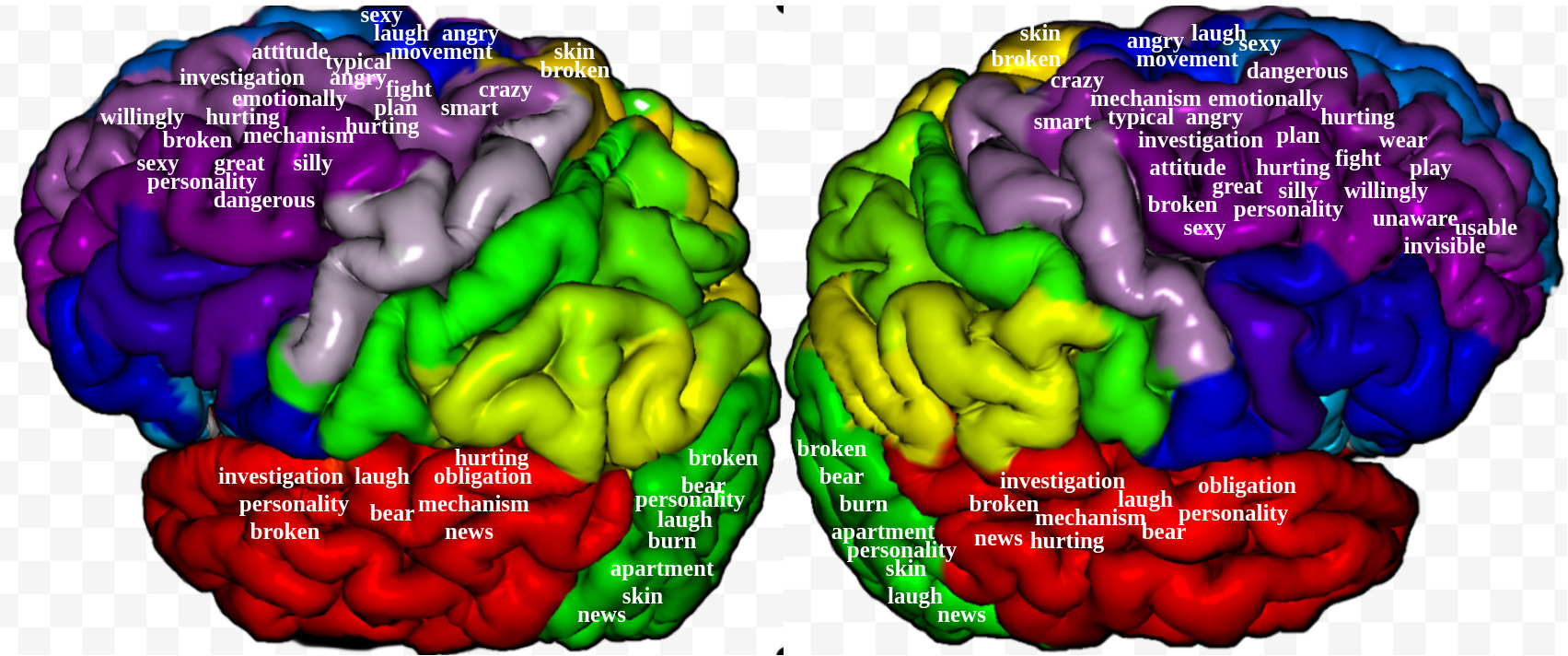}
\endminipage\hfill
\caption{Specialization of Expert-3 for words and the corresponding regions of interest (ROIs) in the brain}
\label{fig:brainroisexp3}
\end{figure*}

\begin{figure*}[t]
\minipage{\textwidth}
\centering
  \includegraphics[width=0.9\linewidth]{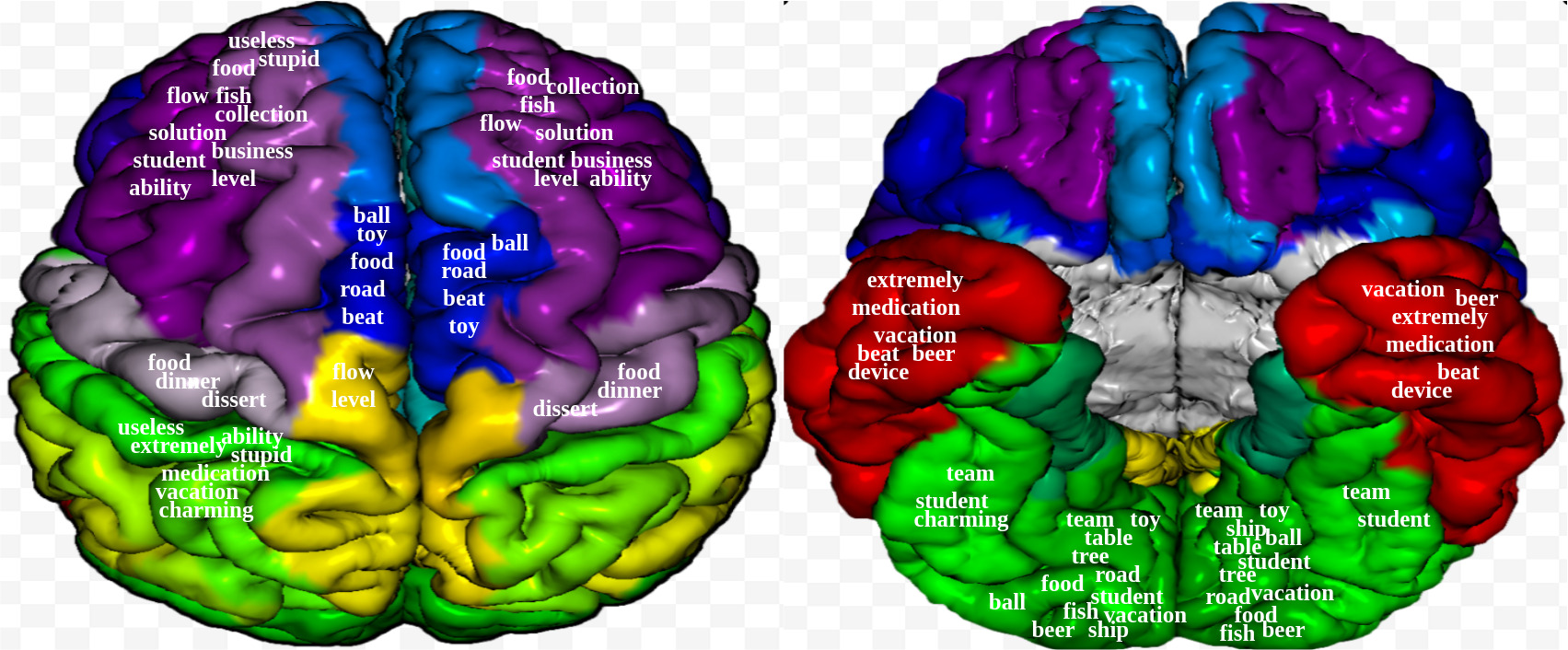}
\endminipage\hfill
\caption{Specialization of Expert-4 for words and the corresponding regions of interest (ROIs) in the brain}
\label{fig:brainroisexp4}
\end{figure*}

\begin{figure*}[t]
\minipage{\textwidth}%
\centering
  \includegraphics[width=0.9\linewidth]{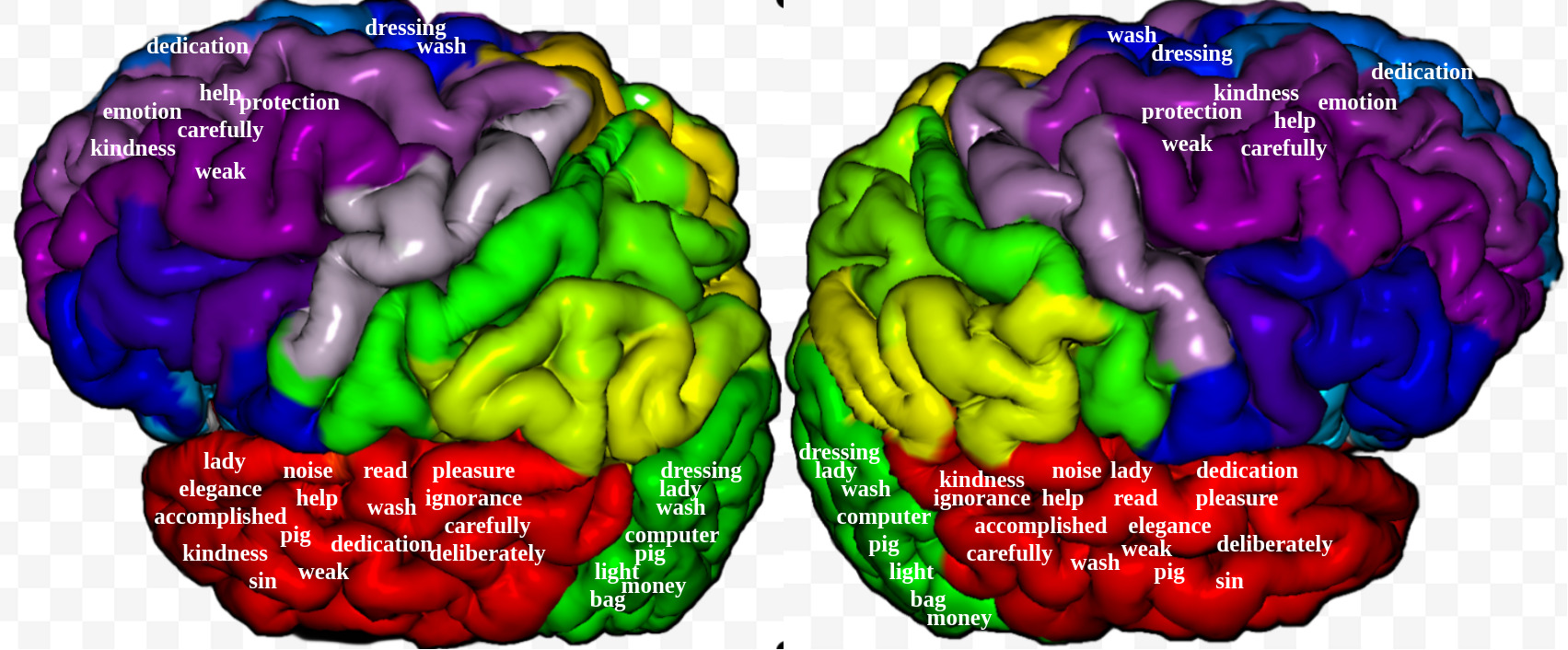}
\endminipage 
\caption{Specialization of Expert-5 for words and the corresponding regions of interest (ROIs) in the brain}
\label{fig:brainroisexp5}
\end{figure*}

It can be observed from Table~\ref{words_results} that the experts correspond to distinct (specialized) ROIs based on joint learning of semantic aspects of the word stimulus and the associated brain regions related to the meaning of the word. 
This aspect can be observed in all the experts, especially in the set of words that have a close correspondence between training and test conditions. 
Further, it has to be noted that the ROIs activated are known from previous studies to have compatibility with the semantics they seem to encode, few of these examples are discussed later. In order to capture all these joint associations in a visually convenient manner, we depict the words and the associated regions of activation on a 3D-rendered depiction on either the left and right cortical surfaces or the dorsal and ventral surfaces of the brain in Figures~\ref{fig:brainroisexp1} $-$ ~\ref{fig:brainroisexp5}. This depiction, the so-called \emph{brain dictionary}, is inspired from the recent work of Huth et al.~\cite{huth2016natural} who suggested that every word activates multiple regions of the brain.

%A recently developed work by~\citep{huth2016natural} describes that every word activates at multiple parts of the brain. Figures~\ref{fig:brainroisexp1},~\ref{fig:brainroisexp2},~\ref{fig:brainroisexp3},~\ref{fig:brainroisexp4},~\ref{fig:brainroisexp5}

%\noindent{\textbf{Expert-1}}
Several brain regions such as ``Angular\_L,R'', ``Lingual\_L,R'', ``Precentral\_L,R'', ``Postcentral\_L,R'', ``Cuneus\_L,R'', ``Frontal\_Sup\_L,R'', ``Frontal\_Mid\_L,R'', ``Precuneus\_L,R'', ``Cerebellum\_Crus\_1L,1R'', ``Temporal\_Sup\_L,R'', and ``Temporal\_Mid\_L,R'' %\textcolor{red}{{\bf LAST ROW IS MISSING NOW, SHOULD WE KEEP THIS IN SUPPLEMENTARY NOW?}}) 
are commonly activated among all the experts (results not shown in~ Table~\ref{words_results}). 
The common ROIs seem to be related generally to visual-spatial processing, sensory processing, attention, etc. that seem to be shared for all the words and may be related to the processing of the visual stimulus presented along with the lexical input (word). Apart from the common activation, there are also unique regions captured by experts in a semantically appropriate fashion. Some of the notable examples are highlighted here. 

From the Figure~\ref{fig:brainroisexp1}, we observe that \textbf{Expert1} seems to code for action words such as ``play", ``spoke", noun-verb co-occurrences such as ``event-spoke", ``do dance", ``dig",  ``do music", ``play movie", visual-spatial related words, including, camera, picture, texture, etc~\cite{pulvermuller2013neurons,houk2007action}.
The cortical areas associated with movement such as the Supplementary Motor Area, Cerebellum, Putamen, Caudate seem to be active in {\textbf{Expert 1}}.

It appears that {\textbf{Expert 3}} %%\textcolor{red}{\bf EXP-1??} 
codes for face recognition words such as ``laugh", ``emotions", ``feelings", ``hurting", ``sexy", and ``angry", problem solving words such as ``investigation", ``mechanism", ``news", etc.  The brain activation in the Fusiform gyrus that lies between the Parahippocampal gyrus and the Lingual gyrus medially seem to be compatible with face processing~\cite{weiner2016anatomical,bogousslavsky1987lingual}.

\iffalse
Cortical areas associated with movement such as the Supplementary Motor Area, Cerebellum, Putamen, Caudate seem to be active in {\textbf{Expert 4}} that seems to code for action words such as ``read", ``spoke", noun-verb co-occurrences such as ``event-spoke", ``do dance", ``dig",  ``applause", ``counting", etc~\cite{pulvermuller2013neurons,houk2007action}.
\fi

{\textbf{Expert-5}} codes for abstract words such as accomplished, pleasure, elegance, dedication, the brain activation in the polar regions of the temporal lobe known for involvement in higher-order language comprehension seems compatible. 
Similar correspondences can be seen in several other words such as read, noise activated in both temporal and frontal area known for hearing and speech functionalities respectively.. 

%Although, as can be seen from Table~\ref{words_results} that some words that the experts seem to code for are unrelated, by and large, MoRE model seems to succeed in capturing associations between word features and relevant brain activation. 

%for are thought to be controlled mostly by the Left Caudate and the Thalamus, which can be observed in Expert 4 for the words ``read", ``spoke" and combinatorial learning of noun-verb co-occurrences ``event-spoke", etc.  [{\bf ANY REFERENCE HERE ?? NEVER HEARD OF CAUDATE IN LANGUAGE OR COMMUNICATION}]
%ROIs associated with visual areas Cerebellum, Left, and Right Occipital Mid area is known for the words such as ``bird" and ``picture" in the same Expert 4. [{\bf SOME CAREFUL RE-LOOKING IS NEEDED. CB IS NOT KNOWN TO BE VISUAL AREA? ANY SUCH STATEMENTS REQUIRE REFERENCE TO BE CITED}]

%Other regions Insula, Right Putamen and Right Caudate seem compatible. 

%Figure~\ref{fig:brainrois} showcases expert-wise specialization in the regions of activation (ROIs) in the brain. 
%Here, we depict slices from the coronal section of the brain, indicating the voxels corresponding to unique brain regions activated by each expert. MoRE models seem to learn the association between brain regions and the semantic meaning of the words. 
%We measured the accuracy using voxel intensities and the location of voxel coordinates between the predicted and actual data. 

\subsection{ROI Prediction for Unseen Words}
\label{unknown_words}

\iffalse
\begin{table}[!htb]
  \caption{MoRE prediction results for Unseen words. The last row lists the common regions activated across all the experts.} 
  \label{Expert}
  \begin{tabular}{c}
      \includegraphics[width=\linewidth]{images/tab3.png} \\
      \includegraphics[width=0.25\linewidth]{images/physics.pdf} \includegraphics[width=0.25\linewidth]{images/Lunch.pdf}
      \includegraphics[width=0.25\linewidth]{images/Generous.pdf}
      \includegraphics[width=0.25\linewidth]{images/Cat.pdf}\\
      \textbf{(a) Physics} \hspace{1.1cm} \textbf{(b) Lunch} \hspace{0.9cm} \textbf{(c) Generous} \hspace{1.1cm} \textbf{(d) Cat} \\
  \end{tabular}
  
\end{table}
\fi

%\setlength{\tabcolsep}{0.5pt}
{\renewcommand{\arraystretch}{0.5}
\begin{table*}[t]
\scriptsize
\centering
\caption{MoRE prediction results for unseen words. The last row lists the common regions activated across all the experts.}
\label{Expert}
\begin{tabular}{|c|c|c| c|} \hline \hline
\textbf{Unseen Words} &\textbf{(Highlighted Expert)} & \textbf{(Related Words)} & \textbf{Brain Regions}\\ \hline \hline
 Physics & Expert-1 & Science,  & ParaHippocampal\_L \\
 & &Mathematics & Fusiform\_L, Supramarginal\_L\\ 
 & & & Occipital\_Mid\_L, Rectus\_L \\ \hline
 Lunch & Expert-4 & Dinner & Supp\_Motor\_Area\\
 & & & Supramarginal\\
 & & &Frontal\_Inf\_Oper \\ \hline
Generous & Expert-5 &  Kindness & Temporal\_Pole\_Mid\\
& & & Temporal\_Pole\_Sup \\\hline
Cat & Expert-1 & dog & Cerebellum\_6\\
& & & Occipital\_Mid\_L,R \\
& & & Cerebellum\_Crus\_1L,1R \\\hline 
\multicolumn{4}{|c|}{Angular\_L,R, Precentral\_L,R, Postcentral\_L,R, Cuneus\_L,R }  \\ 
\multicolumn{4}{|c|}{Frontal\_Sup\_L,R, Frontal\_Mid\_L,R, Precuneus\_L,R}  \\
\multicolumn{4}{|c|}{Cerebellum\_Crus\_1L,1R, Temporal\_Sup\_L,R, Temporal\_Mid\_L,R}  \\\hline
\multicolumn{4}{|c|}{\includegraphics[width=0.25\linewidth]{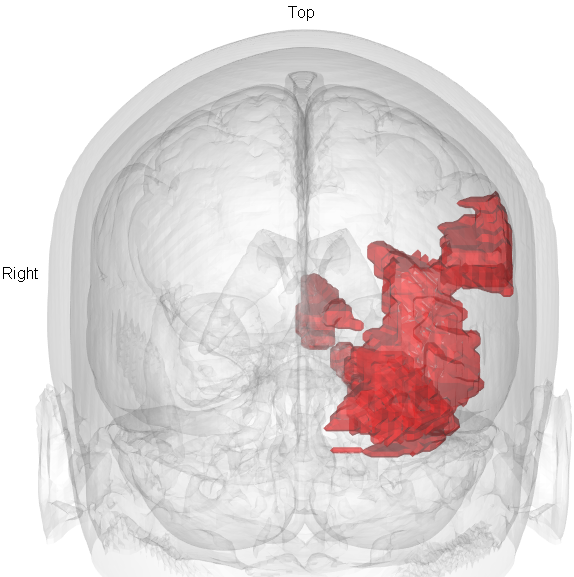}   \includegraphics[width=0.25\linewidth]{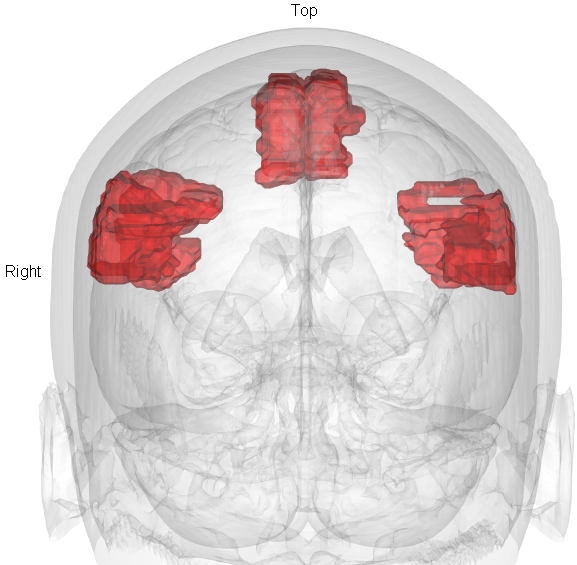} 
      \includegraphics[width=0.25\linewidth]{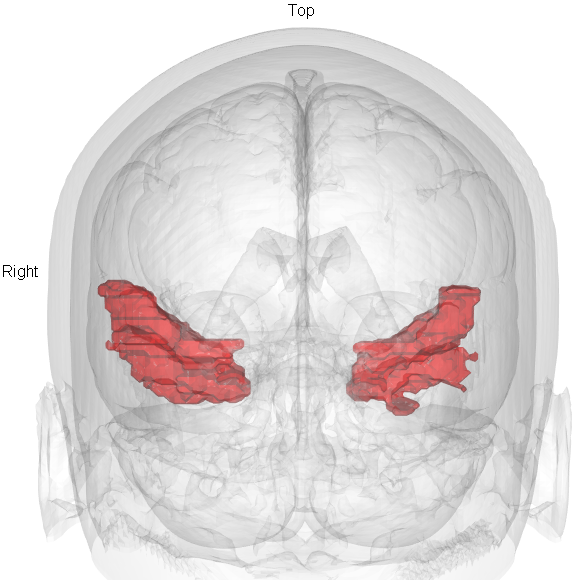} 
      \includegraphics[width=0.25\linewidth]{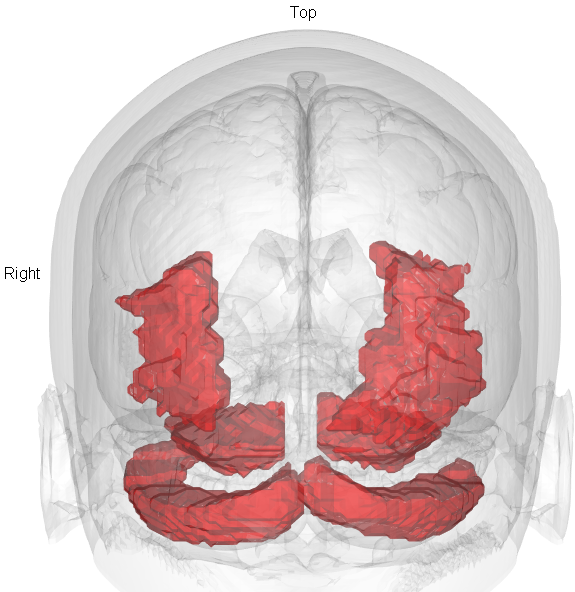}}\\
\multicolumn{4}{|c|}{\textbf{(a) Physics} \hspace{3.2cm} \textbf{(b) Lunch} \hspace{3.2cm} \textbf{(c) Generous} \hspace{3.2cm} \textbf{(d) Cat}}  \\ \hline    
\end{tabular}
\end{table*}}

To measure the efficacy of our proposed MoRE model for the word+picture condition, we tried to predict the brain regions for unseen words, i.e., those that are not present in the dataset. We chose four different words such as ``physics", ``lunch", ``generous", and ``cat" which are semantically related to words in the existing dataset but not explicitly given while training, model prediction results are displayed in Table~\ref{Expert}.  For the word ``physics", MoRE model chose expert-1 with higher probability among the five experts and expert-1 earlier captured the words science and mathematics in training \& testing, respectively as shown in Table ~\ref{words_results}. The results of this unseen-word experiment give credence to our hypothesis that learning functional differentiation (or specialization) while achieving comparable overall accuracy can be implemented with the mixture of regression experts (MoRE) framework.

\section{Conclusion}
\label{sec:conclusion}
In this paper, we present a mixture of experts based model (Expert2Coder) where a group of experts captures brain activity patterns related to particular regions of interest (ROIs) and also shows semantic discrimination across different experts.
Different from previous works, the underlying model depicts that each expert trains on particular brain regions of interest (set of voxels that are significantly activated) based on the semantic category of words that are represented by the model.
%The key distinction of our work is the utilization of experts model (MoRE: Mixture of Regression Experts) to capture the ROIs based on various categories of input. 
Various experiments demonstrated the efficacy and validity of the proposed approach. Notably, the last experiment with words that were not used in training, demonstrates the power of such encoding models that learn a joint association between semantics from linguistic representation and brain responses. These models can potentially predict the brain response corresponding to new words.

In future, we plan to experiment on Spatio-temporal fMRI datasets, with a primary focus on the hierarchical mixture of experts at slice-level instead of voxel-level predictions at each instance.



%\bibliographystyle{IEEEtran}

\bibliographystyle{IEEEtranS}
\bibliography{IEEEabrv}

\section{Supplementary Results}
\label{supplementary}

\begin{figure*}[t]
\begin{center}
  \includegraphics[width=0.8\linewidth]{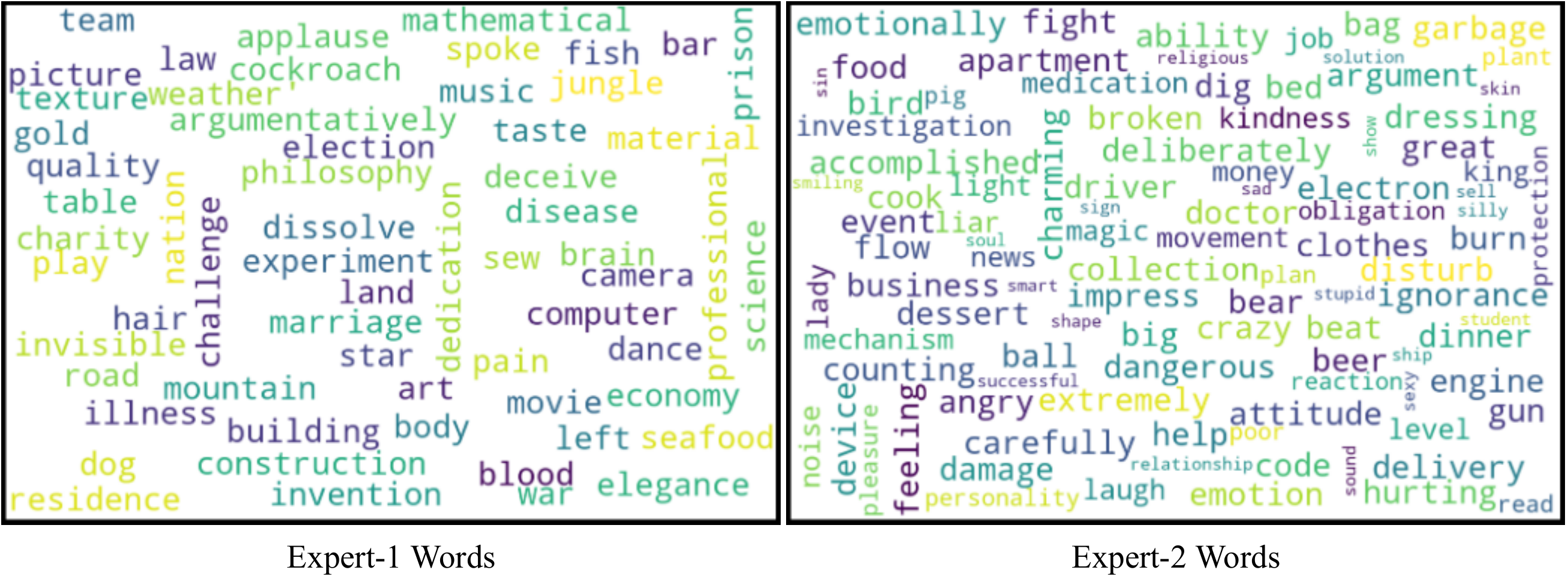}
\caption{Specialization of word categories associated with Brain activations learned by 2-experts in the cross-validation scheme}
\label{fig:fold1brainrois}
\end{center}
\end{figure*}

\begin{figure*}[!htb]
\begin{center}
  \includegraphics[width=0.8\linewidth]{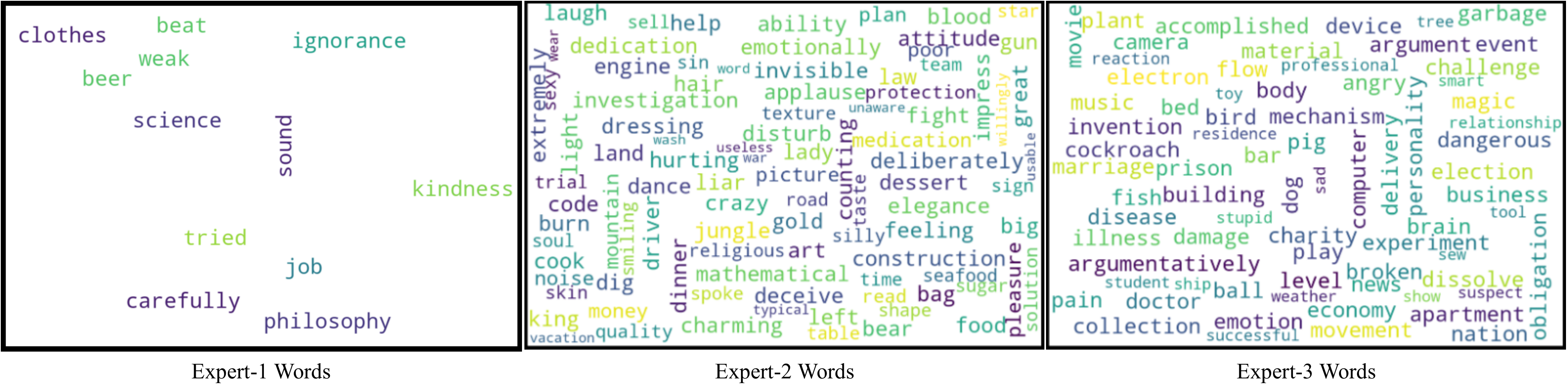}
\caption{Specialization of word categories associated with Brain activations learned by 3-experts in the cross-validation scheme}
\label{fig:fold2brainrois}
\end{center}
\end{figure*}

\begin{figure*}[!htb]
\begin{center}
  \includegraphics[width=0.8\linewidth]{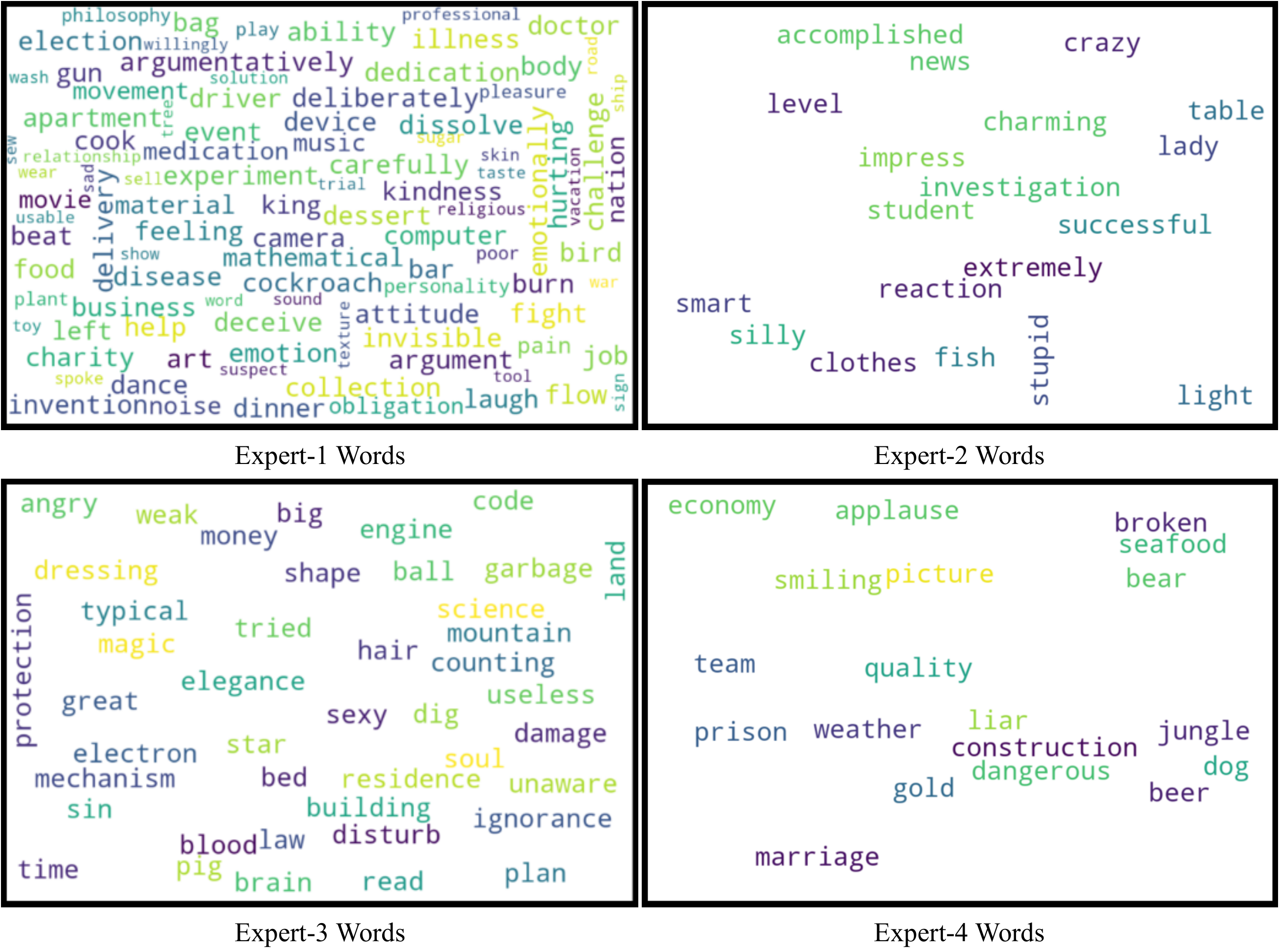}
\caption{Specialization of word categories associated with Brain activations learned by 4-experts in the cross-validation scheme}
\label{fig:fold3brainrois}
\end{center}
\end{figure*}

In this section, we report the MoRE model results when tested with a different number of experts.
As seen in the 2-experts MoRE model results visualization shown in Figure~\ref{fig:fold1brainrois}, some related words tend not to appear together.
Besides, the existence of different categories of words present in 2-experts is not sufficient for learning different ROIs.
Similarly, Figure~\ref{fig:fold2brainrois},~\ref{fig:fold3brainrois} describes the MoRE results with 3-experts and 4-experts.
From Figure~\ref{fig:fold2brainrois}, we observe that expert-1 tries to learn a specific category of words and corresponding brain ROIs.
Also, the observations from Figure~\ref{fig:fold3brainrois} that 3 out of 4 experts categorized the words and learned the unique brain ROIs. 
Overall, we achieved the optimal number as 5-experts discussed in the paper.

\end{document}